\documentclass[preprint,12pt]{elsarticle}

\usepackage{amssymb}
\usepackage{amsthm}
\usepackage{amsmath}
\usepackage[ruled,vlined]{algorithm2e}
\usepackage{geometry}
\newgeometry{left=2cm,right=2cm,top=2cm,bottom=2cm}
 
\usepackage{tabularx}
\usepackage{booktabs}
\usepackage{multirow}

\usepackage{enumitem}
\usepackage{todonotes}
\usepackage{hyperref}

\usepackage{extpfeil}

\usepackage{algorithmic}

\usepackage{tikz}
\newcommand{\circledplus}{%
    \tikz[baseline, scale=0.8]{%
        \draw[draw=black, fill=none] (0,0) circle (1ex); %
        \draw[draw=black, line width=0.5pt] (-1ex,0) -- (1ex,0); %
        \draw[draw=black, line width=0.5pt] (0,-1ex) -- (0,1ex); %
    }%
}

\newcommand{\circledminus}{%
    \tikz[baseline, scale=0.8]{%
        \draw[draw=black, fill=none] (0,0) circle (1ex); %
        \draw[draw=black, line width=0.5pt] (-1ex,0) -- (1ex,0); %
    }%
}

\newcommand{\centeredDash}{\makebox[5em][c]{-}}

\usepackage{caption}
\usepackage{makecell}

\makeatletter
\def\@date{}
\makeatother

\begin{document}

\begin{frontmatter}

\title{Unifying Physics- and Data-Driven Modeling via Novel Causal Spatiotemporal Graph Neural Network for Interpretable Epidemic Forecasting}

\author[FIAS,ITP,XIJFC]{Shuai Han}
\author[FIAS,ITP]{Lukas Stelz}
\author[FIAS]{Thomas R. Sokolowski}
\author[CUHK,FIAS]{Kai Zhou}
\author[FIAS,ITP,GSI]{Horst Stöcker}


\affiliation[FIAS]{
    organization={Frankfurt Institute for Advanced Studies},
    postcode={60438},
    city={Frankfurt am Main},
    country={Germany}
}

\affiliation[ITP]{
    organization={Institut für Theoretische Physik, Goethe Universität Frankfurt},
    postcode={60438},
    city={Frankfurt am Main},
    country={Germany}
}

\affiliation[XIJFC]{
    organization={Xidian-FIAS International Joint Research Center},
    postcode={60438},
    city={Frankfurt am Main},
    country={Germany}
}

\affiliation[CUHK]{
    organization={School of Science and Engineering, The Chinese University of Hong Kong},
    postcode={518172},
    city={Shenzhen},
    country={China}
}

\affiliation[GSI]{
    organization={GSI Helmholtzzentrum für Schwerionenforschung GmbH},
    postcode={64291},
    city={Darmstadt},
    country={Germany}
}

\begin{abstract} 
Accurate epidemic forecasting is crucial for effective disease control and prevention. Traditional compartmental models often struggle in estimating temporally and spatially varying epidemiological parameters, while deep learning models typically overlook the underlying disease transmission dynamics and lack interpretability in the epidemiological context. To address these limitations, we propose a novel Causal Spatiotemporal Graph Neural Network (CSTGNN), a hybrid framework that integrates a Spatio-Contact SIR model with the Graph Neural Networks (GNNs) to capture the spatiotemporal propagation of epidemics across regions. Inter-regional human mobility exhibits continuous and smooth spatiotemporal patterns, leading to adjacent graph structures that share underlying mobility dynamics. To model these dynamics, we employ an adaptive static connectivity graph to represent the stable components of human mobility and utilize a temporal dynamics model to capture fluctuations within these mobility patterns. By integrating the adaptive static connectivity graph with the temporal dynamics graph, we construct a dynamic graph that encapsulates the comprehensive properties of human mobility networks. Additionally, to capture temporal trends and variations in infectious disease spread, we introduce a temporal decomposition model to handle temporal dependence. This model is then integrated with a dynamic graph convolutional network for epidemic forecasting. We validate our model using real-world datasets at the provincial level in China and at the state level in Germany. Extensive studies demonstrate that our method effectively models the spatiotemporal dynamics of infectious diseases, thus providing a valuable tool for epidemic forecasting and intervention strategies. Furthermore, our analysis of the learned parameters provides valuable insights into the disease transmission mechanisms, enhancing the interpretability and practical applicability of our model. \par 

\end{abstract}


\end{frontmatter}

\section{Introduction}
The swift and widespread transmission of COVID-19, as characterized by its high contagion and significant morbidity, has profoundly affected global economies, international trade, healthcare systems, and, most importantly, countless lives \cite{baker2020economic}. This pandemic has presented burdensome obstacles to almost all nations and regions, resulting in the implementation of various intervention tools to curb viral spread, such as lockdowns, the use of protective masks, and vaccination campaigns \cite{hsiang2020covid}. Precise predictions of the pandemic’s trajectory are essential for policymakers to devise effective containment strategies, optimize resource distribution, and strengthen healthcare infrastructure to protect populations. Nevertheless, the progression of an epidemic is inherently intricate, and traditional mathematical models often fail to adequately represent the complex and nonlinear behaviors observed in real-world scenarios. Fortunately, the significant progress in deep learning across disciplines such as image recognition, natural language processing, and big data analytics \cite{goodfellow2016deep, lecun2015deep} has paved the way for its application in epidemic modeling \cite{yang2021machine, covid19dl2022, wang2023applications}, resulting in enhanced predictive accuracy and more reliable forecasts. \par 

A variety of compartmental models, which are based on systems of differential equations, have been developed to simulate the spread of epidemics within diverse populations. Among these, the classical SIR model \cite{Kermack1927} is widely regarded as a fundamental framework. These models effectively capture the dynamics of epidemic progression by delineating the mobility of individuals between different states within the population. These foundational models have been further extended to incorporate real-world transmission characteristics and interruptive intervention measures, such as latent period transmission \cite{Wallinga2006}, vaccination \cite{Nowak1999}, and asymptomatic infections \cite{Fisman2020}. Integrating these additional factors allows researchers to improve the tracking of epidemic spread complexity in real-world settings, thereby enhancing the prediction accuracy. Moreover, due to the pronounced temporal dependence inherent in epidemics, forecasting is treated as a time series prediction task. Traditional time series analysis methods, including ARIMA \cite{Box1976} and Support Vector Regression (SVR) \cite{Smola1998}, have been applied to epidemic predictions with varying degrees of success. In recent years, deep learning models have demonstrated significant advantages in this domain: notable models include Recurrent Neural Networks (RNNs) \cite{Rumelhart1986}, Long Short-Term Memory networks (LSTM) \cite{Hochreiter1997}, Gated Recurrent Units (GRUs) \cite{Cho2014}, Temporal Convolutional Networks (TCNs) \cite{Bai2018}, Physics-Informed Neural Networks (PINNs) \cite{HAN2024106671,Raissi2019}, and attention-based architectures such as the Transformer \cite{Vaswani2017} and the Temporal Fusion Transformer (TFT) \cite{Lim2020}.\par 

The actual spread of infectious diseases is, However, influenced not only by temporal factors, but also by the spatial heterogeneity of populations across different regions, and their fluctuations that may have both deterministic and stochastic components. Traditional models, such as the SIR framework and its variants, typically assume a homogeneous population, thus failing to account for the distinguishing characteristics and interactions of populations in various areas \cite{Smith2020,Johnson2019}. Hence, the simplifying assumption of homogeneity overlooks important key aspects, like population density, mobility patterns, and local intervention measures, leading to inaccurate predictions and ineffective control strategies. To address these limitations, it is essential to incorporate population heterogeneity into epidemiological models, in order to capture various aspects of spatial dependence which affect disease transmission \cite{Garcia2021}. This approach requires integrating both, temporal and spatial dimensions, ensuring that the dynamic progression of an epidemic is also linked to interactions among populations in different, widely separated regions. Considering spatial heterogeneities, alongside temporal trends, new models can improve the accuracy and reflect the complexity of disease spread in diverse environments. Recent research advancements, such as the development of metapopulation models that account for regional interactions \cite{Miller2022,cao2023metapopulation}, the application of network-based approaches to better understand mobility patterns \cite{Davis2023}, and the use of machine learning techniques to integrate spatial data \cite{Lee2023}, have significantly enhanced the predictive accuracy. These improvements ultimately provide more effective information for public health strategies, offering a lead to better-informed and well targeted interventions. \par 

However, notwithstanding these improvements, existing models hardly explicitly account for population mobility, which extends the transmission beyond the confines of individual regions and thereby profoundly influences the spread of epidemics. \cite{kraemer2020effect, wu2020dynamic}. This introduces dynamically both, temporal variability and spatial dependence, into epidemic data. Hence, to rely solely on time-series models for accurate predictions is insufficient \cite{yu2018spatio}. Instead, epidemic forecasting is best approached as a spatiotemporal prediction task: historical data are modeled along both the spatial and temporal dimensions, to uncover latent transmission patterns \cite{cao2020spatial}. Graph Neural Networks (GNNs) have emerged as powerful tools for handling spatial data in non-Euclidean spaces, driving a surge in GNN-based research \cite{wu2020comprehensive, sun2023attention}. These models typically combine time-series methods to capture temporal dependence with graph-based algorithms to simulate spatial relationships \cite{li2018diffusion}. Most existing spatiotemporal models rely on prior knowledge for constructing static graph structures, such as using geographic proximity to create adjacency networks \cite{kraemer2020effect}. However, real-world interregional interactions are far more complex \cite{barabasi2004network}. To address these issues, researchers have developed weighted adjacency matrices based on proximity, incorporating population interaction data as well as integrating multiple sources of prior knowledge—such as adjacency relationships, migration flows, and travel distances—to construct richer relational graphs \cite{yu2018spatio, li2018diffusion}. Gravity-based models, which account for factors like population size and distance, have also been employed to measure interregional influences and to build relational graphs based on time-series similarity \cite{wu2020dynamic}. While these approaches provide valuable insights, the resulting graph structures are often overly simplistic, incomplete, or biased, and often fail to fully capture the complexity of population mobility in epidemic transmission \cite{huang2020adaptively}. To overcome these limitations, adaptive adjacency matrices have been proposed. These matrices leverage learnable embeddings, to uncover dynamically the latent relationships between nodes. Thus, a flexible and comprehensive way is offered to model the intricate spatial dependence in epidemic spread \cite{li2018diffusion}. \par 

Despite significant progress in both mathematical modeling and machine learning, several challenges remain: First, when using compartmental models and their variants for infectious disease modeling, there is often a trade-off between model complexity and data availability. Specifically, as the level of detail in modeling real-world scenarios increases, the number of unknown epidemiological parameters grows correspondingly, such that the existing observational data can not drive effectively such complex models \cite{anderson1992infectious, heesterbeek2015modeling}. Second, in deep learning approaches, particularly those based on GNNs, most methods either independently construct static graphs or adaptive dynamic graphs \cite{wu2020comprehensive, yu2018spatio}. Hence, they often fail to account for the actual population mobility patterns and for the complex interactions between regions which play a critical role in the spread of infectious diseases \cite{kraemer2020effect, wu2020dynamic}. As a result, these models struggle to capture the true dynamics of real-world population movements. Moreover, directly generated dynamic graphs often lead to high computational complexity and costs. This poses challenges for optimization when using standard loss functions \cite{li2018diffusion, sun2023attention}. Hence, solely relying on those traditional methods or simple deep learning technically introduce additional limitations. Traditional methods unfortunately often fail to leverage the substantial advantages of deep learning, such as its ability to learn and encode complex non-linear relationships in a purely data-driven fashion, even when no theoretical framework describing these relationships is available \cite{lecun2015deep}. On the other hand, deep learning models are often hard to interpret, which, however, is crucial for understanding and for addressing the spread of infectious diseases \cite{lipton2018mythos, rudin2019stop}. This dual gap underscores the need for approaches that can integrate the strengths of both, traditional modeling and deep learning, while approaching these issues in a computationally efficient and interpretable manner \cite{karniadakis2021physics, rackauckas2020universal}. \par 

To address the aforementioned challenges, we first developed a Spatio-contact SIR (SCSIR) model that incorporates the spatial dimension into epidemic modeling. By introducing the concept of contact rate, the model quantifies population interactions across different regions. Second, we integrated this model, as causal prior knowledge, into a spatiotemporal graph neural network, thus proposing a novel epidemic forecasting approach called Causal Spatiotemporal Graph Neural Network (CSTGNN). To model a realistic population mobility, we assume that most individuals exhibit a few distinct movement patterns or that groups share a fundamental mobility structure far-near across adjacent time steps. Previous approaches constructed either static or dynamic graphs independently. Here, we design a framework which combines adaptive static graphs with learned dynamic temporal graphs. The adaptive static graph captures shared structural information over a fixed time interval, while the dynamic temporal graph utilizes temporal modeling learns time-dependent graph structures. By integrating these two graph representations, we establish a spatiotemporal dynamic graph to represent the learned spatiotemporal features of nodes. In addition, and inspired by findings in \cite{MAO2024111952}, we have observed that spatiotemporal decomposition models effectively capture both, the trends and the variations of infectious disease transmissions over time. This allows the model to effectively identify both, short-term fluctuations and long-term patterns, in epidemic spread. This can be achieved by employing a time-series decomposition-based model to handle temporal dependence and combining this with Graph Convolutional Networks (GCNs), the learned dynamic graphs address spatial dependence. Thus, comprehensive spatiotemporal epidemic forecasting is enabled:

\begin{enumerate}[label=\textbullet]
    \item A novel epidemiological model is introduced, which incorporates contact rates as a key metric to account for inter-regional population interactions, providing a causal foundation for epidemic modeling.

    \item An appropriate graph structure learning method is realized to effectively simulate the complexity of human mobility patterns in the context of epidemic transmission. Specifically, an adaptive static graph is applied to model stable mobility patterns and to employ a temporal convolution model, to capture variations in mobility dynamics.
    
    \item Causal prior knowledge from the SCSIR model is integrated with human mobility patterns as learned through the graph-based model, designing the present CSTGNN. This new tool explicitly learns both, time- and region-varying epidemiological parameters, as well as latent epidemic propagation patterns across regions, in a fully end-to-end manner from heterogeneous epidemic data.

    \item Extensive computational evaluations are performed on two distinctly different datasets, to evaluate the performance of this CSTGNN. The results demonstrate that CSTGNN achieves competitive to state-of-the-art accuracy for both, short-term and long-term forecasting. Moreover, analyses of the learned parameters yield deep insights into mechanisms of disease transmission. These insights reveal that human mobility plays a significant role in accelerating epidemic spread, thereby enhancing the interpretability and rapid practical utility of the present model.
\end{enumerate}

The remainder of this paper is organized as follows: Section \ref{sec:rw} reviews related work. Section \ref{sec:me} provides a detailed description of the proposed model structure. Section \ref{sec:ex} presents the simulated experimental results and offers an in-depth analysis of the findings. Finally, Section \ref{sec:co} concludes the paper with a summary of our contributions and a discussion of potential directions for future work.

\section{Related Work}
\label{sec:rw}

\subsection{Compartmental Models}
Following the general review of approaches towards the study of epidemic dynamics, the present section provides a detailed introduction to the compartmental models which are employed in our approach. The objective is to elucidate here how these models are used to simulate epidemic spread dynamics and to explain the underlying mathematics.\par

Mathematical modelling of epidemiological dynamics is a well-established area of research in applied mathematics. The conventional compartmental models were developed to model the dynamics of epidemics within a population. A plain, yet powerful well-known compartmental model is the SIR model \cite{peeri2020sars}, from which several models, such as SIRD and SEIR, originated. SIR assumes that the total population $N$ remains constant. $N$ is divided into three separate groups or compartments: susceptible (S), infectious (I) and removed (R), at each time t. Individuals are transferred between compartments as shown in Figure~\ref{fig:sir_model} with certain rates, called transition rates, $\beta S(t) I(t)/N$ and $\gamma I(t)$:
\begin{figure}[ht!]
	\centering
	\includegraphics[width=15cm]{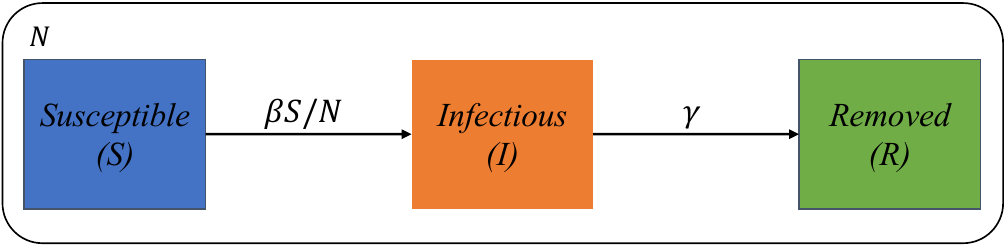}
    \captionsetup{font=normalsize}
    \caption{Schematic illustration of the interactions between compartments in the SIR model.}
	\label{fig:sir_model}
\end{figure}
The simplest epidemiological compartmental models treat all individuals in the same compartment as sharing identical features. Therefore, each compartment is homogeneous. The SIR model is described by the following set of coupled ordinary differential equations:
\begin{equation}
    \begin{aligned}
    \frac{d S(t)}{d t} &=-\frac{\beta I(t)}{N} S(t)\;, \\
    \frac{d I(t)}{d t} &=\frac{\beta I(t)}{N} S(t) -\gamma I(t)\;, \\
    \frac{d R(t)}{d t} &=\gamma I(t)\;.
    \end{aligned}
    \label{equ:sir_eqs}
\end{equation}

Here parameter $\beta$ is the effective transmission rate. It denotes the number of effective contacts made by one infectious and one susceptible individual, which lead to one infection per unit of time. The removal rate $\gamma$ indicates the fraction of infectious individuals who recover or die per unit of time. $\gamma$ can be calculated using 1/D, with D being the average time duration that an infected individual can carry and transmit the virus. Equation~\eqref{equ:sir_eqs} is subject to the the following initial conditions, $S\left(t_{0}\right) > 0, I\left(t_{0}\right) \geq 0, \text { and } R\left(t_{0}\right) \geq 0$ at the initial time $t_{0}$. By construction, $S(t) + I(t) + R(t) = N$ holds at any time $t$. In general, the time scale of the epidemic dynamics is assumed to be short as compared to the length of the life of individuals in the population: the effects of births and natural deaths on the population are therefore not accounted for.\par 

In the classical approach, the following steps are taken for parameter estimation and prediction in the SIR model:
\begin{enumerate}[label=\Roman*]
  \item \textbf{Data Collection}: Gather data on infectious disease cases as observations.
  \item \textbf{Parameter Estimation}: Fit the model to the observed data by tuning $\beta$ and $\gamma$, starting from the reasonable initial values. Minimizing the discrepancy between the model's predictions and the observed data provides the optimal (best fitting) parameter values.
  \item \textbf{Forecasting}: Use these optimized parameter values to predict future epidemic spread dynamics.
\end{enumerate}

\begin{figure}[ht!]
	\centering
	\includegraphics[width=17.5cm]{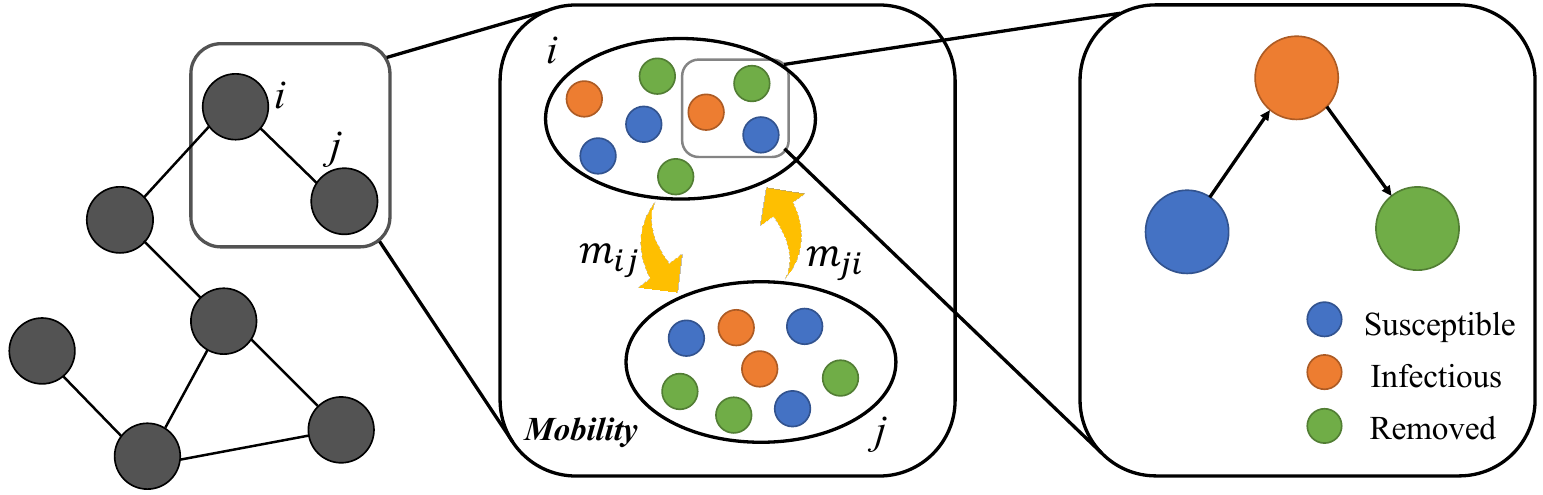}
    \captionsetup{font=normalsize}
	\caption{Illustration of individual homogeneity and heterogeneity. Each black circle represents a distinct region, as shown in the center-left of the figure. The central part of the figure illustrates individual movements between two regions, $i$ and $j$. The right side of the figure depicts individuals transitioning between the $S$, $I$, and $R$ types within a single region.}
	\label{fig:mobility}
\end{figure} \par
The original SIR model is too simple, but it offers a reasonable basic framework for simple simulations of the dynamics of epidemic spread. However, the SIR model, with its oversimplified structure, has significant limitations. In particular, it struggles to account for population diversity and regional variation. In general, differences among individuals—such as age, gender, and other factors—have a significant impact on their behavior. This leads to an important variability in the resulting infection dynamics, known as the population diversity. Diversity is further subjected to regional differences: The simplest SIR model addresses primarily the spread of an epidemic within a single, homogeneous population or region, but neglects the interactions between different regions or sub-populations. In reality, epidemics often exhibit heterogeneous dynamics across various regions. This also involves complex interactions among regions, as illustrated in Figure~\ref{fig:mobility}. To address these limitations, the Spatio-Contact SIR model, as detailed in Equation \eqref{eq:ssir_model}, incorporates regional heterogeneity and human mobility patterns to better capture transmission dynamics across multiple regions. \par
\begin{figure}[ht!]
	\centering
	\includegraphics[width=17.5cm]{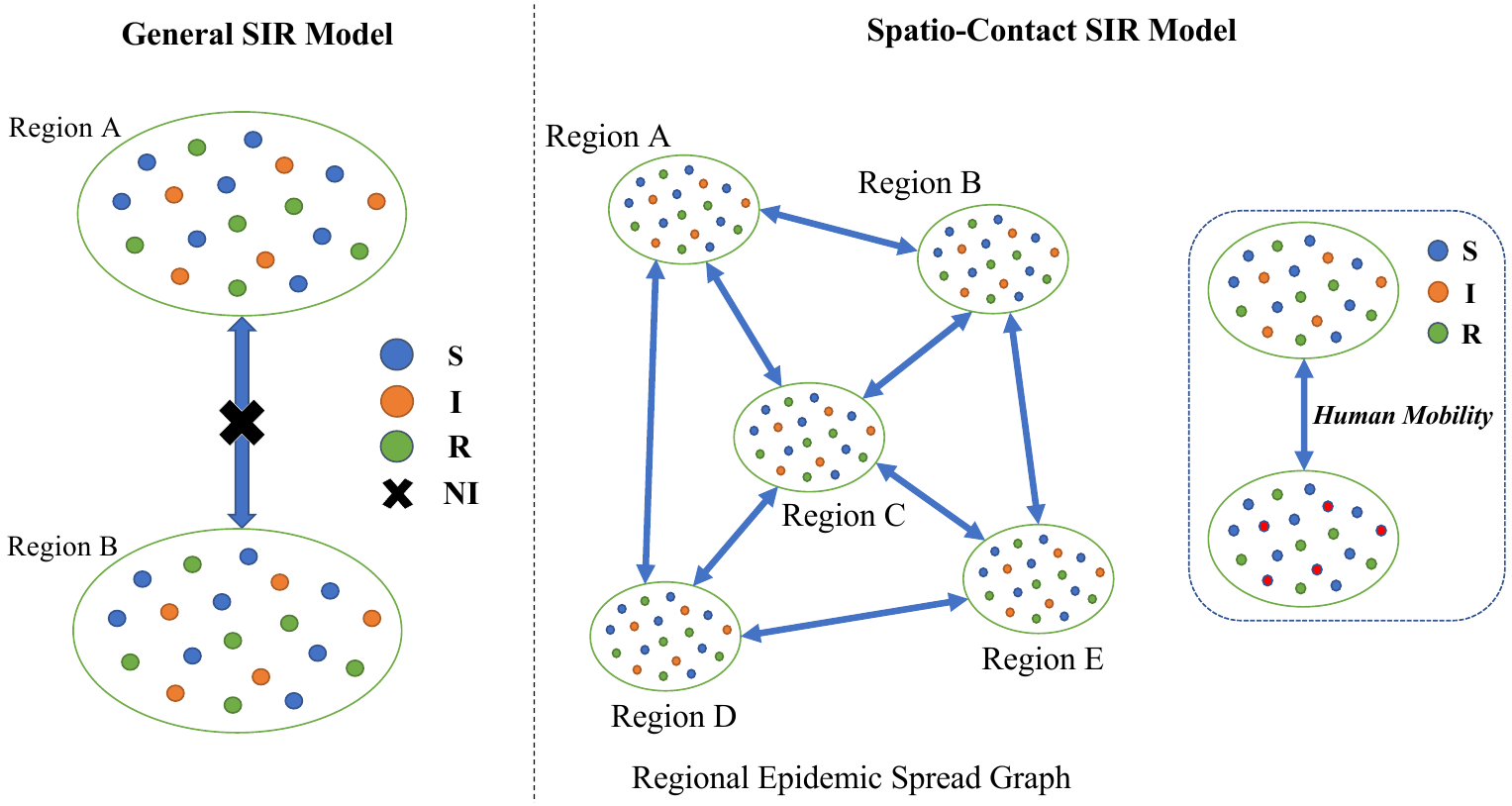}
    \captionsetup{font=normalsize}
	\caption{Difference between the general SIR model and the Spatio-Contact SIR model, NI denotes no interaction between regions.}
	\label{fig:model_comparsion}
\end{figure} \par
The distinction between the SIR model and the Spatio-Contact SIR model is depicted in Figure~\ref{fig:model_comparsion}: While the SIR model tracks transitions between susceptible (S), infectious (I), and recovered (R) individuals within a single region, it does not account for the movement of individuals between different regions.  In contrast, the Spatio-Contact SIR model categorizes the population of each region $i$ into three compartments: $S_i(t)$ for susceptible individuals, $I_i(t)$ for infectious individuals, and $R_i(t)$ for recovered or deceased individuals in region $i$ at time $t$. The total population size of region $i$, denoted by $N_i(t)$, is given by $N_i(t) = S_i(t) + I_i(t) + R_i(t)$. As in the SIR model, $\beta$ represents the infection rate, while $\gamma$ signifies the combined recovery and mortality rate. In addition, the Spatio-Contact SIR model incorporates a contact rate $c_{ij}$ to account for the intensity of epidemic transmission between region $i$ and region $j$. The original Spatio-Contact SIR model is thus described by the following formulae:
\begin{equation}
\begin{aligned}
    \frac{dS_{i}(t)}{dt} &= -\beta \frac{S_{i}(t)}{N_i(t)} \sum_{j=1}^Q c_{ij} I_{j}(t), \\ 
    \frac{dI_{i}(t)}{dt} &= \beta \frac{S_{i}(t)}{N_i(t)} \sum_{j=1}^Q c_{ij} I_{j}(t) - \gamma I_{i}(t), \\
    \frac{dR_{i}(t)}{dt} &= \gamma I_{i}(t).
\end{aligned}
\label{eq:ssir_model}
\end{equation}

Here the parameter $c_{ij}$ is used to form an epidemic transmission graph, as illustrated in the middle of Figure~\ref{fig:mobility}, and $i,j \in \lbrace 1, ..,Q\rbrace$ loops over all regions. Realistic contact rates are in general positively correlated with the mobility rates between regions. Therefore, the Spatio-Contact SIR model provides a more reasonable simulation of the spread of infectious diseases, both within and between regions, by incorporating the contact rate $c_{ij}$, which better reflects the actual transmission dynamics of infectious diseases. \par

\section{Methodology}
\label{sec:me}
In this section, the problem of spatiotemporal epidemic forecasting is defined. The overall framework of the proposed model is outlined and the specific details of each component are discussed.\par

\subsection{Problem Formulation}
This study focuses on forecasting the number of infectious individuals for multiple regions and multiple time steps by using observed epidemic data. \par

Epidemic propagation across different regions can be represented by a graph structure, denoted as $G(\mathcal{V}, \mathcal{E})$. Here $\mathcal{V}$ represents the set of regions and $\mathcal{E}$ denotes the set of weighted edges representing the interactions between these regions. This graph $G$ can be transformed into an adjacency matrix $A \in \mathbb{R}^{Q \times Q}$, where each element $A_{ij}$, for $i, j \in \{1, \ldots, Q\}$, indicates the edge weight between region $i$ and region $j$, capturing the strength of human mobility across these regions. The edge weight reflects the intensity of epidemic transmission across regions which varies over time, to account for the changes of the populations movement. \par 

These temporal dynamics are incorporated by a dynamic adjacency matrix, denoted as $\mathcal{A}_{1:T_{\rm obs}} = [A_1, A_2, \ldots, A_{T_{\rm obs}}] \in \mathbb{R}^{T_{\rm obs} \times Q \times Q}$, representing the evolution of regional connections over $T_{\rm obs}$ equidistant time points. The spatiotemporal features of the epidemic data are then expressed as $\mathcal{X}_{1:T_{\rm obs}} = [X_1, X_2, \ldots, X_{T_{\rm obs}}] \in \mathbb{R}^{T_{\rm obs} \times Q \times F}$. Here each $X_t$ for $t \in \{1,..., T_{\rm obs}\}$ represents the $F$-dimensional historical epidemic data for $Q$ regions at time step $t$. This dataset includes daily counts of the susceptible, infectious, removed (recovered and deceased) cases.\par
\begin{figure}[ht!]
	\centering
	\includegraphics[width=17.5cm]{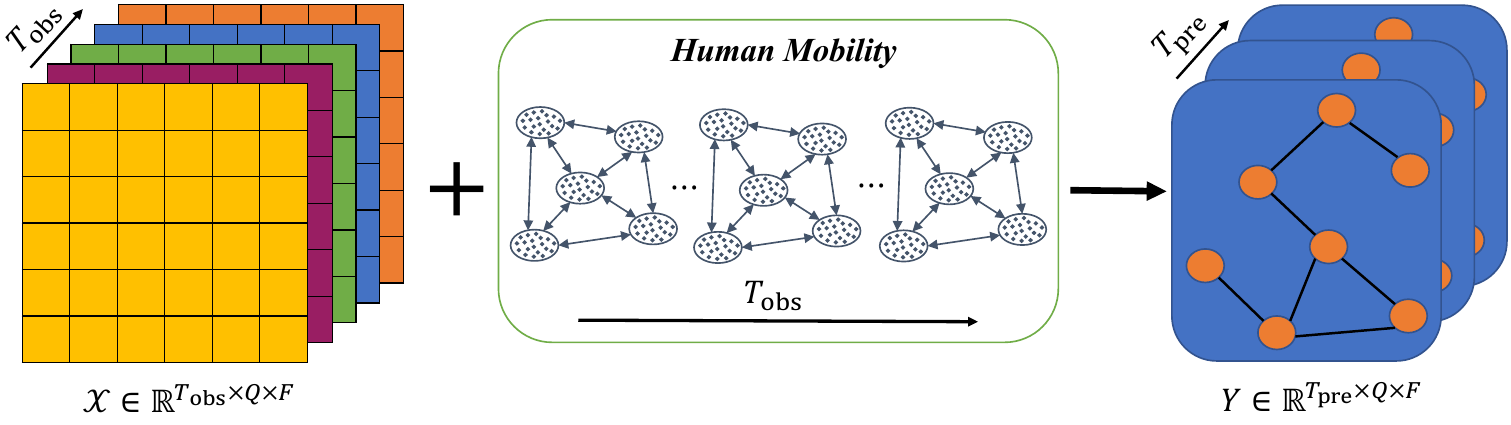}
	\captionsetup{font=normalsize}
    \caption{Illustration of the problem formulation. The observed data $\mathcal{X}$ over $T_{\rm obs}$, combined with the dynamic adjacency matrix $\mathcal{A}_{1:T_{\rm obs}}$, predicts the epidemic states $Y$ over $T_{\rm pre}$, where $T_{\rm obs}$ and $T_{\rm pre}$ denote observed and predicted time horizons, respectively.
}
	\label{fig:problem_formulation}
\end{figure}
The goal of the spatiotemporal epidemic forecasting of this study is to leverage the historical epidemic data $\mathcal{X}$, along with the dynamic adjacency matrix $\mathcal{A}$, to learn a mapping function $f(\cdot)$. This function aims to predict the number of the infectious individuals for the $Q$ regions over the $T_{\rm pre}$ future time steps, covering the period from $T_{\rm obs}+1$ to $T_{\rm obs} + T_{\rm pre}$, denoted as $Y_{T_{\rm pre}} \in \mathbb{R}^{T_{\rm pre} \times Q \times F}$. Thus, in accord with Figure~\ref{fig:problem_formulation}, the epidemic forecasting can be formulated as follows:
\begin{equation}
\left\{\mathcal{X}_{1:T_{\rm obs}},\mathcal{A}_{1:T_{\rm obs}}\right\} \xrightarrow{f(\cdot)} \mathbf{Y}_{T_{\rm obs}+1: T_{\rm obs}+T_{\rm pre}}
\end{equation}

\subsection{Model Overview}
\begin{figure}[ht!]
	\centering
	\includegraphics[width=17.5cm]{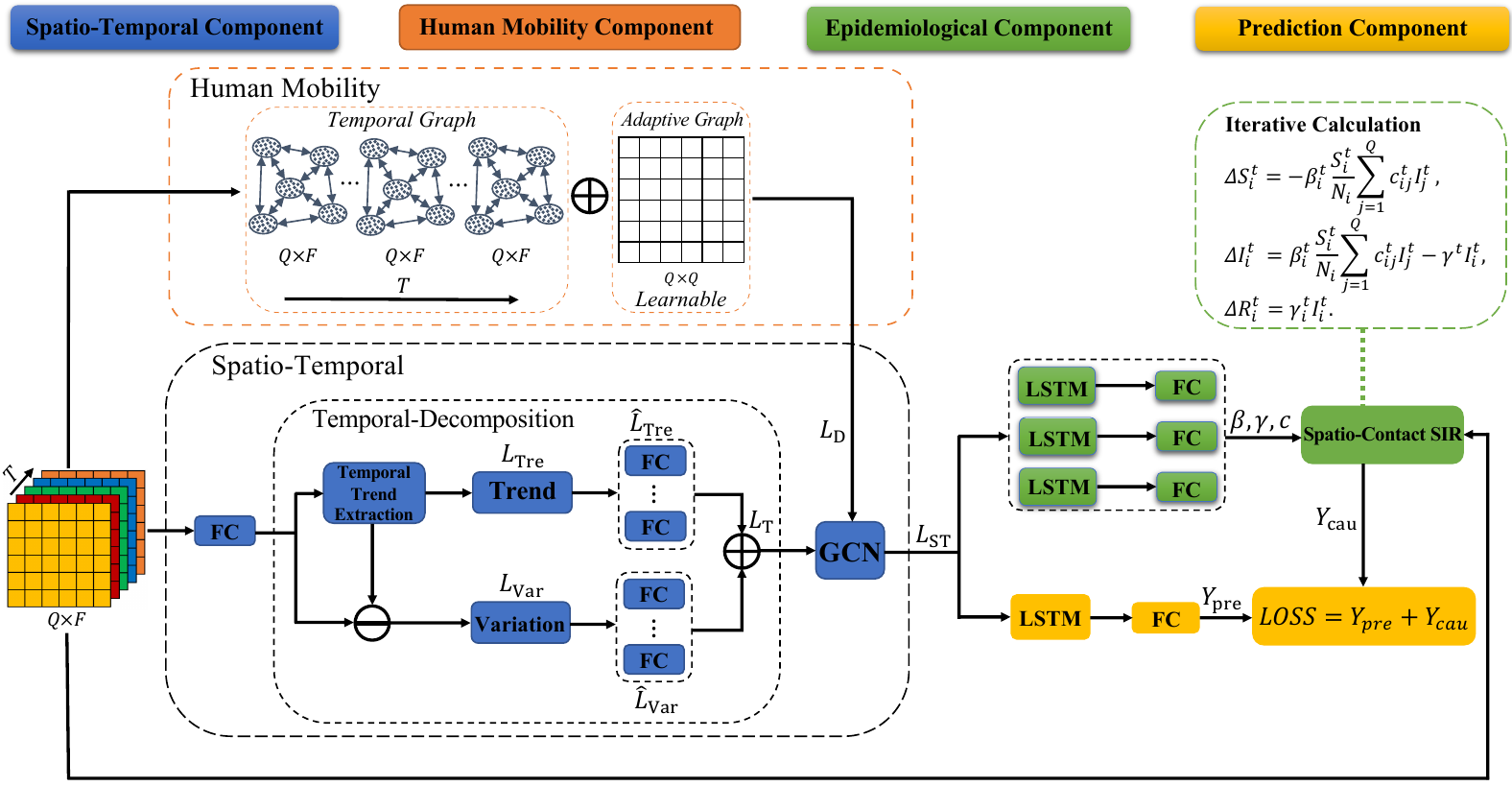}
    \captionsetup{font=normalsize}
	\caption{Illustration of the causal spatiotemporal graph neural network framework.}
	\label{fig:st_e_f}
\end{figure}
The present model is illustrated in Figure~\ref{fig:st_e_f}: Four key components are integrated: the Spatio-Temporal Component, the Human Mobility Component, the Epidemiological Component, and the Prediction Component. The Spatio-Temporal Component employs a time decomposition model and a Graph Convolutional Network (GCN) to capture the spatiotemporal dynamics of epidemic data and predicts parameter sequences for the Spatio-Contact SIR model. The Human Mobility Component learns human mobility patterns between regions, and outputs this information as a dynamic adjacency matrix to the Spatio-Temporal Component. The Epidemiological Component incorporates domain-specific knowledge to refine the model's representation of epidemic dynamics through learned epidemiological parameters. And finally, the Prediction Component combines neural network forecasts with epidemiological insights to jointly constrain predictions, thus enabling more reasonable multi-regional, multi-time-step forecasts of infectious individuals. The various parameters depicted in Figure~\ref{fig:st_e_f} are described in more detail in Table \ref{tab:st_e_f}.
\begin{table}[ht!]
\centering
\captionsetup{font=normalsize}
\caption{Explanation of the parameters in Figure~\ref{fig:st_e_f}.}
\begin{tabular*}{\textwidth}{@{\extracolsep{\fill}} c l}  
\hline
{Symbol} & {Description} \\
\hline
$T, Q, F$ & Temporal, regional, and feature dimensions of epidemic data \\
$L_{\rm Tre}$ & Trend dependence \\
$L_{\rm Var}$ & Variation dependence \\
\circledminus & Element-wise subtraction by dimensional elements \\ 
$\hat{L}_{\rm Tre}$ & Trend dependence after feature-wise transformation \\
$\hat{L}_{\rm Var}$ & Variation dependence after feature-wise transformation \\
\circledplus & Element-wise addition by dimensional elements \\
$L_{\rm T}$ & Temporal dependence \\
$L_{\rm D}$ & Dynamic adjacency dependence \\
$L_{\rm ST}$ & Spatiotemporal dependence \\
$\beta$ & Predicted infection rate \\
$\gamma$ & Predicted recovery rate \\
$c$ & Predicted contact rates\\
$Y_{\rm pre}$ & Neural network prediction \\
$Y_{\rm cau}$ & Epidemiological inference \\
\hline
\end{tabular*}
\label{tab:st_e_f}
\end{table}

\subsection{Spatio-Contact SIR model}
\label{ssir_comp}
Although the SIR model is foundational in epidemiology, its homogeneous mixing assumption limits its capacity to capture regional transmission dynamics. In contrast, the Spatio-Contact SIR model incorporates population heterogeneity and advanced graph learning techniques to more accurately reflect the complex spatiotemporal spread of epidemics. \par

Here, interactions are modeled both within and between regional populations by using the Spatio-Contact SIR model. Accordingly, the contact rate $c_{ij}$ in Equation~ \eqref{eq:ssir_model} is defined to represent the intensity of human mobility between regions. So, higher mobility between regions results in a higher contact rate. The spatiotemporal heterogeneity emerges in the dynamics of epidemic spread due to the influence of factors such as policies, weather, and their dynamic changes over time. To capture these variations, the model allows the infection rate $\beta$, the recovery rate $\gamma$, and the contact rates $c_{ij}$ to vary across both space and time. $\beta$ is stabilized and prevented from becoming excessively small when multiplied by $S_i^t$, we simplify the calculation of $S$ by merging $\beta_i^t$ and $S_i^t$ into a single term on the right-hand side of the equation for $S$. This adjustment assumes a relatively stable $S$ value within each time step, which ensures numerical stability. Based on these adjustments, A discretized extended version of the original Spatio-Contact SIR model can be written down (Equation \eqref{eq:ssir_model}) as follows:
\begin{equation}
\begin{aligned}
    \frac{S_{i}^{t+1} - S_{i}^t}{\Delta t} &= - \frac{\beta_{i}^{t}}{N^t_i} \sum_{j=1}^Q c_{ij}^{t} I_{j}^t, \\ 
    \frac{I_{i}^{t+1} - I_{i}^t}{\Delta t} &= \frac{\beta_{i}^{t}}{N^t_i} \sum_{j=1}^Q c_{ij}^{t} I_{j}^t - \gamma_{i}^{t} I_{i}^t, \\
    \frac{R_{i}^{t+1} - R_{i}^t}{\Delta t} &= \gamma_{i}^{t} I_{i}^t.
\end{aligned}
\label{eq:mssir_model}
\end{equation}

Here, to facilitate iterative calculations, the time is discretized with a constant interval $\Delta t = 1$. With this assumption, the learned epidemic parameters $\beta_i^{t}$, $\gamma_i^{t}$, and $c_{ij}^{t}$ can be used. Here $i,j \in \{1,2,...,Q\}$. The count of the $S$, $I$ and $R$ individuals can be computed iteratively by:
\begin{equation}
\left[S_i^t, I_i^t, R_i^t \right] 
\overset{\text{Equation~\eqref{eq:mssir_model}}}{\underset{\beta_i^{t}, \gamma_i^{t}, c_{ij}^{t}}{\xlongrightarrow{\hspace*{2cm}}}} 
\left[S_i^{t+1}, I_i^{t+1}, R_i^{t+1}\right]
\end{equation}
The specific calculation process based on causal inference is detailed in Section \ref{ecic_comp}. \par

\subsection{Human Mobility Component}
\label{dg_comp}
Human mobility is a fundamental factor driving the regional spread of epidemics and plays a critical role in the transmission dynamics of infectious diseases. Daily mobility patterns are generally assumed to be stable and then can be represented using static graphs. However, they are strongly impacted by exogenous factors such as weather conditions, policy interventions, and social activities. Consequently, static graphs are insufficient to capture the full dynamic, complex nature of epidemic processes. Existing approaches to dynamic graph learning do attempt to address these limitations by generating spatiotemporal graphs which simulate spatiotemporal variations. However, such methods often neglect the explicit modeling of static mobility patterns, as they primarily focus on capturing temporal variations. This bias arises because many dynamic graph learning approaches are designed to emphasize changes over time, often treating static patterns as redundant or implicitly embedded in the evolving structure. However, neglecting static mobility patterns can hinder the seamless integration of static and dynamic components, leading to inconsistencies in the learned representations. \par 

Furthermore, many spatiotemporal graphs lack essential constraints, which results in significant structural discrepancies across successive time steps. These inconsistencies impair the ability of the graphs to reflect the smooth temporal evolution inherent in real-world epidemic processes and exacerbate the challenges of parameter optimization. In addition, these approaches frequently fail to encapsulate the latent mobility patterns that govern disease transmission, thereby limiting their effectiveness in accurately modeling and forecasting the spread of epidemics. Several studies \cite{MAO2024111952,yang2021discrete,ye2022learning} suggest that the structure of dynamic graphs is typically continuous and smooth. Leveraging the similarity between adjacent time steps and incorporating temporal variation can help to generate dynamic graphs which can better reflect a realistic dynamical evolution. \par 
\begin{figure}[ht!]
	\centering
	\includegraphics[width=17.5cm]{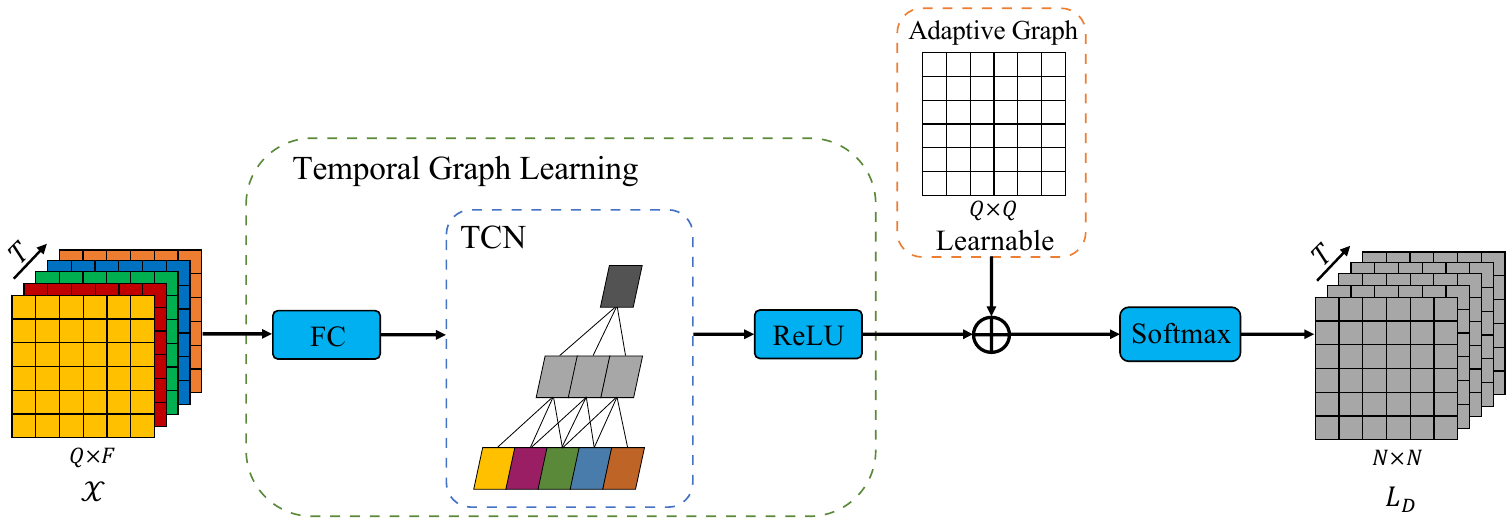}
    \captionsetup{font=normalsize}
    \caption{Illustration of the spatio-temporal graph construction process. Input data is first processed by a fully connected (FC) layer, followed by a Temporal Convolutional Network (TCN) for temporal feature extraction. A ReLU activation is applied, and an adaptive graph refines the representation before the final Softmax output.}
	\label{fig:dgl}
\end{figure}
Inspired by these observations, we construct an adaptive static graph to capture the relatively stable patterns of human mobility, thus reflecting the smoosh nature of most daily activities. Simultaneously, we learn temporal graphs from epidemic data in order to model dynamic variations of population movement over time. We integrate these two components, by introducing constraints into the spatiotemporal graph learning process. This enables the model to better capture both, stable and dynamic aspects of human mobility. This hybrid approach enhances the learning of spatiotemporal graphs, providing an accurate representation of real-world population dynamics during the spatiotemporal evolution of epidemics. The model's structure is shown in Figure~\ref{fig:dgl}. \par

Adaptive learning graph structures have been widely used in the field of graph learning and can provide accurate static graph information. Based on this perspective, we utilize the trainable embedding matrix $A_{\rm static} \in \mathbb{R}^{Q \times Q}$, where $Q$ denotes the number of regions. The dynamic changes of a graph over time, can best be learned by mapping the epidemic data $\mathcal{X} \in \mathbb{R}^{T \times Q \times F}$ to a high-dimensional time-series embedding $A \in \mathbb{R}^{T \times Q \times F_{\rm T}}$, using a fully connected layer. Here $F$ and $F_{\rm T}$ represent the dimensions before and after the mapping:
\begin{equation}
\begin{aligned}
    A &= FC(\mathcal{X}),
\end{aligned}
\end{equation}
this embedding is then fed into a temporal model, to learn the temporal features. It is also passed through the spatio-temporal module for further processing. \par

Temporal Convolutional Networks (TCNs) are widely recognized for their ability to capture long-range temporal dependence in sequential data. By leveraging causal and dilated convolutions, TCNs ensure that the output at each time step depends only on the current and the previous steps, preserving temporal causality. In addition, TCNs support parallel computation and enable efficient training, thus making them particularly suitable for time-series data.

Given the embedding $A \in \mathbb{R}^{T \times Q \times F_{\rm T}}$, TCN processes $A$ along the temporal dimension $T$. The operation in a single TCN layer can be expressed as:
\begin{equation}
\begin{aligned}
A_{\rm temporal}(t, q, f) = \sum_{i=0}^{K-1} \sum_{f'=1}^{F_{\rm T}} W_{i, f', f} \cdot A(t - d \cdot i, q, f') + b_{f}
\end{aligned}
\end{equation}
Here $W \in \mathbb{R}^{K \times F_{\rm T} \times F_{\rm TCN}}$ is the trainable convolution kernel, and $b \in \mathbb{R}^{ F_{\rm TCN}}$ is the bias term. Here, $f$ indexes the output feature channels after the convolution operation, $K$ is the kernel size, and $d$ represents the dilation rate. $F_{\rm T}$ and $F_{\rm TCN}$ denote the input and output feature dimensions, respectively. \par 

By stacking multiple TCN layers, the final output, $A_{\rm temporal} \in \mathbb{R}^{T \times Q \times F_{\rm TCN}}$, is obtained. This representation captures effectively both short-term and long-term temporal dependence, and facilitates downstream spatio-temporal modeling. \par

To construct a hybrid graph which combines both static and dynamic information, the adaptive static graph is merged with the learned temporal graph. To better integrate the static graph $A_{\rm static}$ and the dynamic graph $A_{\rm temporal}$, a fully connected (FC) layer to $A_{\rm temporal}$ is applied. This does not only transform $A_{\rm temporal}$ into a form that matches the shape and dimensions of $A_{\rm static}$, but it also enhances the integration of temporal features into the static structure. Next, an element-wise addition of the static graph $A_{\rm static}$ and the transformed temporal graph $FC(A_{\rm temporal})$ is performed. Numerical stability and interpretability is ensured by applying the $Softmax$ function to the fused graph for normalization, thus mapping its values to the range from 0 to 1. The resulting dynamic graph $L_{\rm D} \in \mathbb{R}^{T \times Q \times Q}$ therefore represents the interaction weights between regions over time. The steps for the calculation are as follows:
\begin{equation}
\begin{aligned}
    L_{\rm D} &= {Softmax}\left( A_{\rm static} \oplus FC\left( A_{\rm temporal} \right) \right),
\end{aligned}
\end{equation}
where $\oplus$ represents element-wise addition. \par

\subsection{Spatio-Temporal Component}
\label{st_comp}
The Spatio-Temporal Component is designed to jointly capture the temporal and spatial propagation dynamics of epidemics. For the temporal modeling module, a time-decomposition approach is adopted, which is widely used in time series forecasting research. That approach focuses on decomposing epidemic data into trends and variations along the temporal dimension. In this way, error accumulation is effectively mitigated and robust learning of temporal dependence is ensured. Spatial features are modeled using a GCN widely used in graph learning research, and integrated with the learned temporal graph in Section \ref{dg_comp}. Combining the extracted temporal and spatial features, this framework provides a comprehensive understanding of both, the temporal evolution and spatial interactions in epidemic data. Hence, a reliable basis for capturing the spatiotemporal spread of infectious diseases is provided. \par

To begin, the mapping of epidemic data $\mathcal{X} \in \mathbb{R}^{T \times Q \times F}$ to a high-dimensional time-series embedding $L \in \mathbb{R}^{T \times Q \times F_{\rm T}}$ is used , employing a fully connected layer. Here $F$ and $F_{\rm T}$ represent the dimensions before and after the mapping,
\begin{equation}
\begin{aligned}
    L &= FC(\mathcal{X})
\end{aligned}
\end{equation}

The Temporal-Decomposition Module is designed to decompose epidemic data along the temporal dimension into two distinct features, Trend and Variation. This process begins with the Temporal Trend Extraction Block, which identifies the overarching temporal trends present in the input data. The trend component $L_{\rm Tre} \in \mathbb{R}^{T \times Q \times F_{\rm T}}$ captures the smooth and long-term evolution of the data. This is computed using a convolution operation with a special kernel, which is designed with fixed and equal weights that sum to 1. This ensures that all values within the temporal window contribute equally to the output. Thus, a smooth and consistent representation of the underlying trend is provided. The variation component $L_{\rm Var} \in \mathbb{R}^{T \times Q \times F_{\rm T}}$ isolates short-term fluctuations from dynamic patterns. It is calculated as the element-wise difference between $L_{\rm Tre}$ and the input data. Both, $L_{\rm Tre}$ and $L_{\rm Var}$, are then processed through fully connected layers to learn refined representations. Thus outputs $\hat{L}_{\rm Tre} \in \mathbb{R}^{T \times Q \times F_{\rm T}}$ and $\hat{L}_{\rm Var} \in \mathbb{R}^{T \times Q \times F_{\rm T}}$ retain their respective temporal features. Finally, the reconstructed output $L_{\rm T}$ integrates these learned representations. This ensures that the decomposed components effectively describe the temporal dynamics of the epidemic data. This decomposition process not only reduces error accumulation, but also enhances the robustness of downstream modeling tasks. The corresponding calculation formulas are
\begin{equation}
\begin{aligned}
    (L_{\rm Tre}, L_{\rm Var}) &= TemporalDecomposition(L), \\
    L_{\rm Tre} &= MovingAverage(Padding(L)), \\
    L_{\rm Var} &= L \mathbin{\circleddash} L_{\rm Tre}
\end{aligned}
\end{equation}
Here $TemporalDecomposition()$ refers to the process of decomposing the time series into its trend and variation components. $MovingAverage()$ represents the moving average operation, applied with padding, to handle boundary effects. This ensures that all time steps are included in the computation. $\ominus$ represents element-wise subtraction.

The trend $L_{\rm Tre}$ and variation $L_{\rm Var}$ components, extracted through temporal decomposition, are processed by fully connected layers, applied separately to each feature along the feature dimension. These operations produce updated temporal embedding representations, $\hat{L}_{\rm Tre}$ and $\hat{L}_{\rm Var}$, corresponding to the trend and variation components, respectively. Finally, the updated trend and variation components are aggregated through an addition operation to produce the final embedding representation, $L_{\rm T}$, which effectively captures temporal dependence. The computation process can be described as follows:
\begin{equation}
\begin{aligned}
    \hat{L}_{\rm Tre} &= FC_{feature}(L_{\rm Tre}), \\
    \hat{L}_{\rm Var} &= FC_{feature}(L_{\rm Var}), \\
    L_{\rm T} &= \hat{L}_{\rm Tre} \mathbin{\oplus} \hat{L}_{\rm Var}.
\end{aligned}
\end{equation}

In addition to temporal dependence, the spread of epidemics is significantly influenced by spatial interactions among regions. To incorporate this spatial impact, the temporal embedding $L_{\rm T} \in \mathbb{R}^{T \times Q \times F_{\rm T}}$ and the dynamic adjacency graph $L_{\rm D} \in \mathbb{R}^{T \times Q \times \rm Q}$ are utilized, which are learned according to Section \ref{dg_comp}, as inputs to the $GCN$ model. The $GCN$ aggregates spatiotemporal information by modeling the interactions between regions through the adjacency relationships encoded in $L_{\rm D}$ and by the temporal features in $L_{\rm T}$. Specifically, the dynamic graph $L_{\rm D}$, which represents normalized spatial connections, is applied to $L_{\rm T}$ in order to propagate information across connected regions. This propagation is followed by a linear transformation parameterized by $W$. This method generates an embedding representation, $L_{\rm ST} \in \mathbb{R}^{T \times Q \times F_{\rm T}}$, effectively capturing both temporal and spatial dependence. The computation process is:
\begin{equation}
\begin{aligned}
L_{\rm ST} &= GCN(L_{\rm D}, L_{\rm T}) \\
&= ReLU(L_{\rm D} L_{\rm T} W + b)
\end{aligned}
\end{equation}
Here $W \in \mathbb{R}^{T \times Q \times F_{\rm T}}$ represents the weight parameters, $b$ denotes the bias term, and $ReLU$ is the activation function. This framework allows the $GCN$ to learn how spatial interactions and temporal dependence jointly influence the spread of epidemics. This makes the resulting representation $L_{\rm ST}$ well-suited for downstream predictive tasks. \par

\subsection{Epidemiological Component}
\label{ecic_comp}
Among epidemic researches, there is a growing consensus that relying solely on spatiotemporal models is insufficient for higher accurately forecasting infectious disease spreading dynamics. These models typically capture only the patterns and distributions inherent in the data, often lacking the explicit incorporation of fundamental physical rules which govern epidemic dynamics. Given the extensive array of domain knowledge models available in epidemiological research, integrating these models into neural networks as prior knowledge could enhance the networks' ability to better grasp the underlying physical principles of epidemic dynamics. Such integration promises to render deep learning-based epidemic predictions better interpretable and more robust. \par

This component can integrate epidemiological context by incorporating the causal-based differential equations from the Spatio-Contact SIR model, as detailed in Section \ref{ssir_comp}, into the neural network framework: In the Spatio-Contact SIR model, $N_{i}$ denotes the population of individuals in region $i$, each of whom can be in one of the following states: $S, I, R$. Compartmental models operate under a homogeneous mixing assumption, i.e., every infected individual can directly infect any other individual. The dynamics of epidemic spread in region $i$ at time $t$ are described by the following equations:
\begin{equation}
\begin{aligned}
\Delta S_{i}^{t} &= -\Delta t \cdot \beta^{t}_{i} \frac{S^{t}_{i}}{N^{t}_i} \sum_{j=1}^Q c^{t}_{ij} I^t_{j}, \\ 
\Delta I_{i}^{t} &= \Delta t \cdot \left(\beta^{t}_{i} \frac{S^{t}_{i}}{N^{t}_i} \sum_{j=1}^Q c^{t}_{ij} I^t_{j} - \gamma^{t}_{i} I^t_{i}\right), \\
\Delta R_{i}^{t} &= \Delta t \cdot \gamma^{t}_{i} I^t_{i}
\end{aligned}
\label{eq:delta_ssir}
\end{equation}
Here $S^t_i$, $I^t_i$, and $R^t_i$ denote the number of individuals in each state within region $i$ at time $t$, with the constraint $S^t_i + I^t_i + R^t_i = N^t_i$. The parameters $\beta^{t}_i$, $\gamma^{t}_i$, and $c^{t}_{ij}$ represent the infection rate, the removal rate, and the contact rate between regions $i$ and $j$, respectively. In this framework, these parameters are considered unknown causal factors across $Q$ regions at time $t$, and they are inferred using a neural network. The term $\Delta$ refers to the newly added number of individuals in each state. \par
 
By estimating these causal parameters using a neural network, the Spatio-Contact SIR model into the neural network framework can be integrated. Specifically, the spatiotemporal dependence of epidemic spread, as captured by the component $L_{\rm ST} \in \mathbb{R}^{T_{\rm obs} \times N \times F}$, is put into a neural network. This neural network is designed to learn both the infection rates, removal rates, and contact rates $\beta, \gamma \in \mathbb{R}^{T_{\rm pre} \times N \times 1}$, $c \in \mathbb{R}^{T_{\rm pre} \times N \times N}$, which govern epidemic dynamics. These rates are then normalized using the Sigmoid function to ensure they remain within reasonable bounds.
\begin{equation}
\begin{aligned}
    \beta = Sigmoid(FC(LSTM(Reshape(L_{\rm ST})))), \\
    \gamma = Sigmoid(FC(LSTM(Reshape(L_{\rm ST})))), \\
    c = Sigmoid(FC(LSTM(Reshape(L_{\rm ST}))))
\label{eq:epi_params_learned}
\end{aligned}
\end{equation}
Here $Sigmoid()$ refers to the activation function, $LSTM()$ represents the Long Short-Term Memory network (LSTM), which is used to capture temporal dependence in the data and to make predictions for future time steps. $FC()$ denotes the fully connected layer in the neural network, and $Reshape()$ changes the shape of the data. \par

Finally, the outbreak’s evolution across all regions can be predicted by combining the last observed epidemic data $(S_{T_{\rm obs}}, I_{T_{\rm obs}}, R_{T_{\rm obs}})$ with the estimated parameters $\beta$, $\gamma$, and $c$ from the Spatio-Contact SIR model and by applying Equation~\eqref{eq:delta_ssir}:
\begin{equation}
\begin{aligned}
(\Delta S_{1:T_{\rm pre}}^{\rm cau},\Delta I_{1:T_{\rm pre}}^{\rm cau},\Delta R_{1:T_{\rm pre}1}^{\rm cau}) &= SCSIR((S_{T_{\rm obs}:-1},I_{T_{\rm obs}:-1},R_{T_{\rm obs}:-1}),\beta,\gamma,c)
\end{aligned}
\end{equation}
Here $SCSIR$ refers to the Spatio-Contact SIR model, and $(\Delta S_{1:T_{\rm pre}}^{\rm cau}, \Delta I_{1:T_{\rm pre}}^{\rm cau}, \Delta R_{1:T_{\rm pre}}^{\rm cau}) \in \mathbb{R}^{T_{\rm pre} \times N \times 1}$ denote the changes in the numbers of susceptible, infectious, and removed individuals across all regions over $T_{\rm pre}$ time steps.

\subsection{Prediction Component}
As illustrated in Figure~\ref{fig:st_e_f}, the entire framework is constrained by two outputs: neural network output and causal inference output. The prediction from the neural network output is used as our final prediction, as it embeds the hidden information from all components.\par
The output of the neural network is generated by feeding the feature $L_{\rm ST}$, as learned by the Spatio-Temporal Component, which captures the dynamics of the epidemic spread, into a fully connected layer with an activation function. This approach enables prediction of the epidemic's temporal evolution, specifically the number of infectious individuals $Y_{\rm pre}$, across all regions for the subsequent $T_{\rm pre}$ time steps:
\begin{equation}
Y_{1:T_{\rm pre}}^{\rm pre} = ReLU(FC(LSTM(Reshape(L_{\rm ST}))))
\end{equation}
Here $FC()$ and $LSTM()$ serve the same functions as introduced in \eqref{eq:epi_params_learned}, and $ReLU()$ denotes the activation function. \par
The output of the causal inference is obtained by iteratively applying Equation~\ref{eq:delta_ssir}. This process utilizes the estimated causal parameters to run the Spatio-Contact SIR model. Thus, the causal prediction for the number of infectious individuals is generated across all regions for the next $T_{\rm pre}$ time steps:
\begin{equation}
\begin{aligned}
\relax (S_{1}^{\rm cau},I_{1}^{\rm cau},R_{1}^{\rm cau}) &= (S_{T_{\rm obs}:-1},I_{T_{\rm obs}:-1},R_{T_{\rm obs}:-1}) + (\Delta S_{1}^{\rm cau},\Delta I_{1}^{\rm cau},\Delta R_{1}^{\rm cau}),\\
(S_{2}^{\rm cau},I_{2}^{\rm cau},R_{2}^{\rm cau}) &= (S_{1}^{\rm cau},I_{1}^{\rm cau},R_{1}^{\rm cau}) + (\Delta S_{2}^{\rm cau},\Delta I_{2}^{\rm cau},\Delta R_{2}^{\rm cau}),\\
&\phantom{=} \vdots \\
(S_{T_{\rm pre}}^{\rm cau},I_{T_{\rm pre}}^{\rm cau},R_{T_{\rm pre}}^{\rm cau}) &= (S_{T_{\rm pre}-1}^{\rm cau},I_{T_{\rm pre}-1}^{\rm cau},R_{T_{\rm pre}-1}^{\rm cau})+(\Delta S_{T_{\rm pre}}^{\rm cau},\Delta I_{T_{\rm pre}}^{\rm cau},\Delta R_{T_{\rm pre}}^{\rm cau}),\\
\end{aligned}
\end{equation} 
Here $(S_{1:T_{\rm pre}}^{\rm cau},I_{1:T_{\rm pre}}^{\rm cau},R_{1:T_{\rm pre}}^{\rm cau}) \in \mathbb{R}^{T_{\rm pre} \times N \times 1}$ represents the causal prediction for each state at the $T_{\rm pre}$ time steps, and $Y_{1:T_{\rm pre}}^{\rm cau} \in \mathbb{R}^{T_{\rm pre} \times N \times 1 }$ ($=I_{1:T_{\rm pre}}^{\rm cau}$) indicates the projected number of infectious individuals for the following $T_{\rm pre}$ time steps, in accord with the Spatio-Contact SIR model. \par

\subsection{Optimization}
The proposed approach is optimized by minimizing the discrepancy between the predicted quantities and the ground truth by using a loss function based on the Mean Absolute Error (MAE), formulated as follows:
\begin{equation}
L(\theta) = \frac{1}{Q \times T} \sum_{q=1}^{Q} \sum_{t=1}^{T} \left( |Y_{q,t}^{\rm pre} - Y_{q,t}^{\rm obs}| + |Y_{q,t}^{\rm cau} - Y_{q,t}^{\rm obs}| \right)
\end{equation}
Here $\theta$ denotes the unknown trainable parameters, $Q$ denotes the total number of regions, $T$ represents the number of predicted future time steps, and $Y^{\rm pre}$ and $Y^{\rm cau}$ are the outputs of the forecasting component and the epidemic inference component, respectively. $Y^{\rm obs}$ is the ground truth. This loss function constrains the training process by minimizing the difference between the outputs of both, the forecasting component $Y^{\rm pre}$ and the epidemic inference component $Y^{\rm cau}$, from the ground truth $Y^{\rm obs}$. The average errors across all regions are calculated and prediction time steps. It provides a comprehensive measure of the prediction accuracy, for both components. \par

\section{Experiments}
\label{sec:ex}
\subsection{Datasets}
We collected two distinct real-world datasets related to epidemics for our experiments: the China-Provinces dataset, which includes data from 31 selected Chinese provinces, and the Germany-States dataset, covering all 16 German federal states. These datasets contain cumulative counts of infectious cases, recovered cases, and deceased cases over time, from which the number of susceptible cases can be derived. 
\begin{table}[ht!]
\centering
\small
\captionsetup{font=normalsize}
\caption{Statistical information of each dataset.}
\begin{tabular}{@{}cccccccl@{}}
\toprule
 Dataset &Timeframe &Size &Max &Min &Mean &Std Dev  \\ \midrule
 China-Provinces  &2021.10.30$\sim$2022.10.30  &$365\times31\times3$ &126009300  &0  &15158910  &27561640  \\
 Germany-States  &2021.10.30$\sim$2022.10.30  &$365\times16\times3$  &17175680  &560  &1763944  &2848876  \\ \bottomrule
 \label{tab:datasets_desc}
\end{tabular}
\end{table}   
Statistical information about the datasets is presented in Table~\ref{tab:datasets_desc}, further details are provided below
\begin{enumerate}[label=\textbullet]
  \item \textbf{China-Provinces}: These 31 province-level datasets were collected from DXY, which provides data from the Chinese Center for Disease Control and Prevention (CCDC). These datasets contain province-level population data and COVID-19-related records in China, covering the period from October 30, 2021, to October 30, 2022 (365 days).
  
  \item \textbf{Germany-States}: These 16 state-level datasets were collected by the Robert Koch Institute (RKI) and include population data and COVID-19-related records for each of the 16 states of Germany over the period from October 30, 2021, to October 30, 2022 (365 days).
\end{enumerate} \par
These two datasets thus span an entire year for each of the two countries. They cover various seasons and all major holidays. This comprehensive date range enables to capture seasonal variations and human mobility, both of which significantly impact the spread of epidemics. In these experiments, infectious cases are used as the primary feature, susceptible and removed cases serve as auxiliary features.

\subsection{Comparison Models}
Several baseline mechanistic models are implemented and compared with the proposed model. The performance of the proposed model in predicting the epidemic spread for the 31 provinces in China and for the 16 states in Germany is evaluated.

\begin{enumerate}[label=(\arabic*)]
  \item \textbf{SIR \cite{stgcn2018}}: The SIR model as one of the most fundamental compartmental models in epidemiology accounts for weekly periodicity, if the optimized values of the infection rate $\beta$ and removal rate $\gamma$ from the previous week is used to generate the predictions.
  
  \item \textbf{SCSIR}: The Spatio-Contact SIR model considers regional heterogeneity, and it models interactions between regions. For each region, the optimized $\beta$, $\gamma$, and contact matrix $c$ from the previous week is used to produce the predictions.
  
  \item \textbf{STGCN}: The STGCN structure \cite{das2023geometric} represents one of the pioneering approaches that integrate GCNs and TCNs for spatio-temporal predictions. In this study, a modified version of STGCN is employed, where the fixed adjacency matrix in the GCN is replaced with a trainable graph structure, and the epidemic causal inference module is excluded. All other components remain consistent with our proposed model, to ensure a fair and rigorous comparative analysis.

  \item \textbf{STGODE}: The STGODE framework \cite{Fang2021Spatial} combines graph neural networks with an ordinary differential equation (ODE)-based approach to construct a spatio-temporal graph ODE network. This facilitates continuous-time modeling and prediction. Here, a tailored version of STGODE is used: the fixed adjacency matrix in the GCN is replaced with a trainable graph structure, and the prediction module is excluded. All other components are kept aligned with our proposed model. This ensures a robust and equitable comparative evaluation.
\end{enumerate}

\subsection{Setup of Experiments}
We divide each dataset into training, validation, and test sets, with ratios of $60\%-20\%-20\%$. All data is  normalized to the range $(0, 1)$. The input time length to 7 days is set for evaluation of the performance of the proposed model for both short-term and long-term forecasting. The output time lengths are 7 and 14 days for short-term forecasting, and 21 and 28 days for long-term forecasting, respectively. Both, the temporal-decomposition and the 3-layer GCN, are configured with dimension 16, and denoted as $L_{\rm T}$ and $L_{\rm ST}$. A 1-layer TCN with a kernel size of 3 is used in the graph learning component. During training, the curriculum learning strategy proposed in \cite{Bucci2023} is employed. Here, the prediction horizon is incrementally extended days by day, based on an early stopping criterion, with a maximum of 200 epochs for each horizon. Training starts with a one-day-ahead prediction and progresses iteratively until the full output time length is reached. The model is optimized using Adam's optimizer with a learning rate of $1 \times 10^{-4}$. All experiments are implemented in PyTorch and conducted on two NVIDIA A100 GPUs. 

\subsection{Evaluation Metrics}
Several metrics are employed to evaluate the performance of the proposed model: Mean Absolute Error (MAE), Root Mean Squared Error (RMSE), Relative Absolute Error (RAE), Pearson Correlation Coefficient (PCC), and Concordance Correlation Coefficient (CCC). These metrics assess both, the accuracy of the predictions (MAE, RMSE, RAE) and the degree of agreement between the predicted and observed values (PCC, CCC). Low values of MAE, RMSE, and RAE indicate higher prediction accuracy, while higher values of PCC and CCC suggest a stronger correlation with the ground truth. To mitigate the influence of randomness, we conduct five independent experiments for each model and report the mean values along with the corresponding $95\%$ confidence intervals. The formal definitions of these evaluation metrics are provided here. \par 

MAE quantifies the average magnitude of the prediction errors, disregarding their direction:
\begin{equation}
{MAE} = \frac{1}{Q \times T} \sum_{q=1}^{Q} \sum_{t=1}^{T} \left| Y_{q,t}^{{\rm pre}} - Y_{q,t}^{{\rm obs}} \right|
\end{equation}
Here $Y_{q,t}^{\rm pre}$ and $Y_{q,t}^{\rm obs}$ represent the predicted and observed values at location $q$ and time $t$, respectively.

RMSE is the square root of the average squared differences between predictions and observations, emphasizing larger errors:
\begin{equation}
{RMSE} = \sqrt{ \frac{1}{Q \times T} \sum_{q=1}^{Q} \sum_{t=1}^{T} \left( Y_{q,t}^{{\rm pre}} - Y_{q,t}^{{\rm obs}} \right)^2 }.
\end{equation}

RAE measures the relative magnitude of the total absolute error compared to the total variation of the observed data:
\begin{equation}
{RAE} = \frac{ \sum_{q=1}^{Q} \sum_{t=1}^{T} \left| Y_{q,t}^{{\rm pre}} - Y_{q,t}^{{\rm obs}} \right| }{ \sum_{q=1}^{Q} \sum_{t=1}^{T} \left| Y_{q,t}^{{\rm obs}} - \bar{Y}^{{\rm obs}} \right| }
\end{equation}
Here $\bar{Y}^{\rm obs}$ is the mean of the observed values.

PCC assesses the linear correlation between predicted and observed values:
\begin{equation}
{PCC} = \frac{ \sum_{q=1}^{Q} \sum_{t=1}^{T} \left( Y_{q,t}^{{\rm pre}} - \bar{Y}^{{\rm pre}} \right) \left( Y_{q,t}^{{\rm obs}} - \bar{Y}^{{\rm obs}} \right) }{ \sqrt{ \sum_{q=1}^{Q} \sum_{t=1}^{T} \left( Y_{q,t}^{{\rm pre}} - \bar{Y}^{{\rm pre}} \right)^2 } \sqrt{ \sum_{q=1}^{Q} \sum_{t=1}^{T} \left( Y_{q,t}^{{\rm obs}} - \bar{Y}^{{\rm obs}} \right)^2 } }
\end{equation}
Here $\bar{Y}^{\rm pre}$ and $\bar{Y}^{\rm obs}$ are the means of the predicted and observed values, respectively.

CCC extends PCC by considering both precision and accuracy, measuring the agreement between predicted and observed values:
\begin{equation}
{CCC} = \frac{2 \,\rho \,\sigma_{{\rm pre}} \,\sigma_{{\rm obs}} }{ \sigma_{{\rm pre}}^2 + \sigma_{{\rm obs}}^2 + \left( \mu_{{\rm pre}} - \mu_{{\rm obs}} \right)^2 }
\end{equation}
Herein $\rho$ is the Pearson Correlation Coefficient as calculated above, while $\sigma_{{\rm pre}}$ and $\sigma_{{\rm obs}}$ are the standard deviations, and $\mu_{{\rm pre}}$ and $\mu_{{\rm obs}}$ are the means of the predicted and observed values, respectively. Unlike PCC, CCC accounts for both the correlation and the mean squared difference between the predictions and observations, providing a comprehensive measure of agreement. The CCC ranges from $-1$ to $+1$, where $+1$ indicates perfect concordance and $-1$ indicates total discordance.\par 

Employing these metrics thoroughly evaluates the performance of all models in our subsequent comparisons from multiple perspectives, which ensures a robust assessment of their predictive capabilities.

\subsection{Performance Evaluation}
This experiment compares the CSTGNN model with the four baseline models for short-term and long-term epidemic forecasting. The results of the comparisons are presented in Tables \ref{table:tbpe4china} and \ref{table:tbpe4germany}. Here, boldfaces indicate the best forecasting performance, underlinings represent the second-best (suboptimal) forecasting performance, and the "Improvement" item denotes the enhancement of CSTGNN over the suboptimal results (calculated by ignoring the error components). Symbol "-" indicates that CSTGNN did not achieve the optimal forecasting in that instance. The CSTGNN model generally achieves the best or the most competitive performance, across different forecasting tasks and datasets, as compared to all other models. \par 

\begin{table}[ht!]
\renewcommand{\arraystretch}{1.25}
\setlength{\tabcolsep}{3pt} 
\captionsetup{font=normalsize}
\caption{Performance comparison with baseline models on the China dataset.}
\label{table:tbpe4china}
\large
\resizebox{1.0\textwidth}{!}{
\begin{tabular}{@{}lllllllllll@{}}
\toprule
\multirow{3}{*}{} & \multicolumn{10}{l}{The China dataset} \\ \cmidrule(lr{0.1pt}){2-11} 
                  & \multicolumn{5}{l}{L=7} & \multicolumn{5}{l}{L=14} \\ 
                    \cmidrule(lr{0.1pt}){2-6} \cmidrule(lr{0.1pt}){7-11} 
Model                  & MAE & RMSE & RAE & PCC & CCC & MAE & RMSE & RAE & PCC & CCC \\ \midrule
SIR                    &77.1±1.0 &250.2±7.5 &0.4±0.0 &89.6±0.8\% &75.8±1.8\%
                       &102.8±0.8 &340.8±5.5 &0.6±0.0 &75.5±1.5\% &53.7±1.8\% \\ 
SCSIR                  &79.5±1.1 &261.2±7.5 &0.4±0.0 &88.9±1.1\% &72.9±1.8\%
                       &104.5±0.5 &349.0±2.8 &0.6±0.0 &74.2±0.7\% &50.7±1.0\% \\
STGCN                  &53.2±6.3 &186.6±12.9 &0.5±0.0 &88.2±0.6\% &77.9±0.7\%
                       &92.0±18.7 &155.1±51.8 &0.5±0.1 &94.3±3.9\% &94.0±3.9\% \\
STGODE                 &\underline{44.7±9.9} &\underline{101.6±28.0} &\underline{0.3±0.1} &\underline{97.5±1.5\%} &\underline{96.9±1.7\%}
                       &\underline{65.1±5.0} &\textbf{143.6±11.7} &\underline{0.4±0.0} &\underline{95.4±0.8\%} &\textbf{94.5±1.1\%} \\ 
CSTGNN                 &\textbf{39.6±5.3} &\textbf{83.5±12.1} &\textbf{0.2±0.0} &\textbf{98.3±0.5\%} &\textbf{98.1±0.5\%}
                       &\textbf{58.5±10.3} &\underline{147.7±27.2} &\textbf{0.3±0.1} &\textbf{95.4±1.3\%} &\underline{93.8±3.6\%} \\ 
Improvement            &11.31\% &17.81\%  &12.0\% &0.75\% &1.25\%  
                       &10.16\% &\centeredDash  &8.57\% &0.05\% &\centeredDash \\ \midrule
\multirow{2}{*}{} & \multicolumn{5}{l}{L=21} & \multicolumn{5}{l}{L=28} \\ 
                    \cmidrule(lr{0.1pt}){2-6} \cmidrule(lr{0.1pt}){7-11} 
Model                  & MAE & RMSE & RAE & PCC & CCC & MAE & RMSE & RAE & PCC & CCC \\ \midrule
SIR                    &110.6±1.1 &330.1±4.8 &0.6±0.0 &71.0±1.0\% &49.3±2.2\%
                       &\underline{114.5±1.0} &324.7±4.9 &0.7±0.0 &68.9±1.4\% &47.9±2.0\% \\ 
SCSIR                  &111.6±0.8 &334.4±6.7 &0.6±0.0 &69.9±2.1\% &47.5±2.6\%
                       &116.1±0.5 &333.2±2.2 &0.7±0.0 &67.1±0.5\% &44.0±1.2\% \\
STGCN                  &102.6±31.3 &271.2±91.8 &0.7±0.2 &76.0±12.5\% &73.8±14.8\%
                       &127.5±46.8 &324.0±157.4 &0.7±0.8 &\textbf{78.3±11.3\%} &55.2±13.9\% \\
STGODE                 &\underline{100.1±14.3} &\underline{243.6±54.5} &\underline{0.6±0.1} &\underline{84.4±7.3\%} &\underline{82.8±7.3\%}
                       &127.6±43.2 &\underline{322.4±146.8} &\textbf{0.7±0.3} &\underline{75.8±14.3\%} &\underline{72.3±15.9\%} \\
CSTGNN                 &\textbf{91.0±16.6} &\textbf{213.1±31.8} &\textbf{0.5±0.1} &\textbf{88.3±4.7\%} &\textbf{84.7±7.7\%}
                       &\textbf{111.7±44.1} &\textbf{270.9±100.9} &\underline{0.6±0.3} &75.0±23.8\% &\textbf{73.5±23.0\%} \\ 
Improvement            &9.14\% &12.53\% &10.53\% &4.62\% &2.31\%  
                       &2.46\%	&15.98\% &\centeredDash &\centeredDash &1.69\% \\ \midrule
\end{tabular}
}
\end{table}

On the China dataset, shown in Table \ref{table:tbpe4china}, the CSTGNN model achieves the best performance across all tasks. In short-term forecasting (L = 7, 14), the CSTGNN model demonstrates significant improvements, by achieving at least 11.31\% improvement in MAE, 12.5\% in RMSE, and 0.05\% in PCC, as compared to the second-best performing model (STGODE). Although the CSTGNN model does not achieve an improvement in some metrics, such as CCC for L = 14, it still provides competitive results, ranking second, and with only a small gap as compared to the STGODE model performing best in these cases. This demonstrates that CSTGNN is highly effective in short-term forecasting tasks. In long-term forecasting (L = 21, 28), the CSTGNN model continues to outperform all other models. It achieves at least a 9\% improvement in MAE, 12.53\% in RMSE, and nearly 2\% in CCC; only at L=28 does it trail slightly behind STGCN in PCC and STGCDE in RAE, both by a small edge. Overall, this highlights the CSTGNN's robustness and consistency in handling long-term prediction tasks. \par 

\begin{table}[ht!]
\renewcommand{\arraystretch}{1.25}
\setlength{\tabcolsep}{3pt} 
\captionsetup{font=normalsize}
\caption{Performance comparison with baseline models on the Germany dataset.}
\label{table:tbpe4germany}
\large
\resizebox{1.0\textwidth}{!}{
\begin{tabular}{@{}lllllllllll@{}}
\toprule
\multirow{3}{*}{} & \multicolumn{10}{l}{The Germany dataset} \\ \cmidrule(lr{0.1pt}){2-11} 
                  & \multicolumn{5}{l}{L=7} & \multicolumn{5}{l}{L=14} \\ 
                    \cmidrule(lr{0.1pt}){2-6} \cmidrule(lr{0.1pt}){7-11} 
Model                  & MAE & RMSE & RAE & PCC & CCC & MAE & RMSE & RAE & PCC & CCC \\ \midrule
SIR                    &8527.7±44.2 &15264.1±84.7 &0.5±0.0 &90.5±0.2\% &75.3±0.3\%
                       &10233.8±76.6 &18135.4±168.5 &0.6±0.0 &83.6±0.4\% &62.1±0.9\% \\ 
SCSIR                  &8531.0±65.0 &15276.6±152.8 &0.5±0.0 &90.4±0.3\% &75.3±0.6\%
                       &10286.4±13.1 &18170.2±72.6 &0.6±0.0 &83.6±0.3\% &61.9±0.4\% \\
STGCN                  &7447.2±0.0 &13069.1±0.0 &0.4±0.0 &90.9±0.0\% &82.2±0.0\%
                       &8951.3±0.0 &15923.7±0.0 &0.5±0.0 &84.6±0.0\% &70.8±0.0\% \\
STGODE                 &\underline{5055.0±393.2} &\underline{8308.9±323.0} &\underline{0.3±0.1} &\underline{97.2±1.9\%} &\underline{92.1±4.3\%}
                       &\underline{8208.1±667.2} &\underline{13846.8±469.6} &\underline{0.5±0.1} &\underline{92.2±5.4\%} &\underline{75.5±16.0\%} \\
CSTGNN                 &\textbf{4393.8±416.8} &\textbf{7511.6±184.2} &\textbf{0.2±0.1} &\textbf{97.7±1.3\%} &\textbf{94.4±2.5\%}
                       &\textbf{7400.7±949.6} &\textbf{12577.1±361.6} &\textbf{0.4±0.1} &\textbf{92.7±1.9\%} &\textbf{80.8±11.1\%} \\
Improvement            &9.1\% &12.5\% &10.5\% &4.6\% &2.3\%  
                       &12.5\% &16.0\% &12.3\% &1.0\% &1.7\% \\ \midrule
\multirow{2}{*}{} & \multicolumn{5}{l}{L=21} & \multicolumn{5}{l}{L=28} \\ 
                    \cmidrule(lr{0.1pt}){2-6} \cmidrule(lr{0.1pt}){7-11} 
Model                  & MAE & RMSE & RAE & PCC & CCC & MAE & RMSE & RAE & PCC & CCC \\ \midrule
SIR                    &11605.3±42.3 &20122.5±73.3 &0.7±0.0 &80.1±0.2\% &52.6±0.4\%
                       &11991.0±41.6 &20614.2±62.7 &0.7±0.0 &78.3±0.2\% &47.9±0.4\% \\ 
SCSIR                  &11614.2±66.8 &20119.3±131.0 &0.7±0.0 &80.2±0.3\% &52.5±0.7\%
                       &12008.1±74.2 &20593.2±110.2 &0.7±0.0 &78.4±0.3\% &48.1±0.6\% \\
STGCN                  &10300.5±0.0 &18028.2±0.0 &0.6±0.0 &80.6±0.0\% &61.7±0.0\%
                       &10769.5±0.0 &18735.5±0.0 &0.6±0.0 &78.1±0.0\% &56.7±0.0\% \\
STGODE                 &\underline{9229.0±473.7} &\underline{14772.3±633.3} &\underline{0.5±0.1} &\underline{87.9±2.3\%} &\underline{80.6±8.1\%}
                       &\underline{8460.7±823.0} &\underline{15159.5±664.3} &\textbf{0.5±0.1} &\textbf{87.5±3.2\%} &\underline{71.3±10.6\%} \\ 
CSTGNN                 &\textbf{7487.5±425.9} &\textbf{12220.5±931.9} &\textbf{0.4±0.1} &\textbf{89.0±3.4\%} &\textbf{86.0±4.1\%}
                       &\textbf{8423.9±422.2} &\textbf{13880.9±1406.2} &\underline{0.5±0.1} &\underline{86.2±3.7\%} &\textbf{77.0±15.7\%} \\ 
Improvement            &18.9\%	&17.3\% &19.2\% &1.2\% &6.6\% 
                       &0.4\% &8.4\% &\centeredDash &\centeredDash &8.0\% \\ \midrule
\end{tabular}
}
\end{table}

On the Germany dataset, illustrated in Table \ref{table:tbpe4germany}, the CSTGNN model delivers either the best or highly competitive performance across all forecasting horizons. In short-term forecasting (L = 7, 14), the model achieves at least 9.1\% improvement in MAE, 12.5\% in RMSE, and 2.3\% in CCC, as compared to the second-best model (again STGODE). Although it does not achieve an improvement in some metrics, such as PCC and RAE for L = 28, the CSTGNN model remains close to the STGODE that performs best in these cases and is ranked second, demonstrating its competitiveness in short-term scenarios. In long-term forecasting (L = 21, 28), the CSTGNN model achieves the best performance across all tasks, with at least 18.9\% improvement in MAE, 17.3\% in RMSE, and 8.0\% in CCC. These results further support the CSTGNN model’s superiority and stability in handling long-term prediction tasks. \par

\begin{figure}[ht!]
    \centering
    \begin{minipage}[b]{0.498\textwidth}
        \centering
        \includegraphics[width=\textwidth]{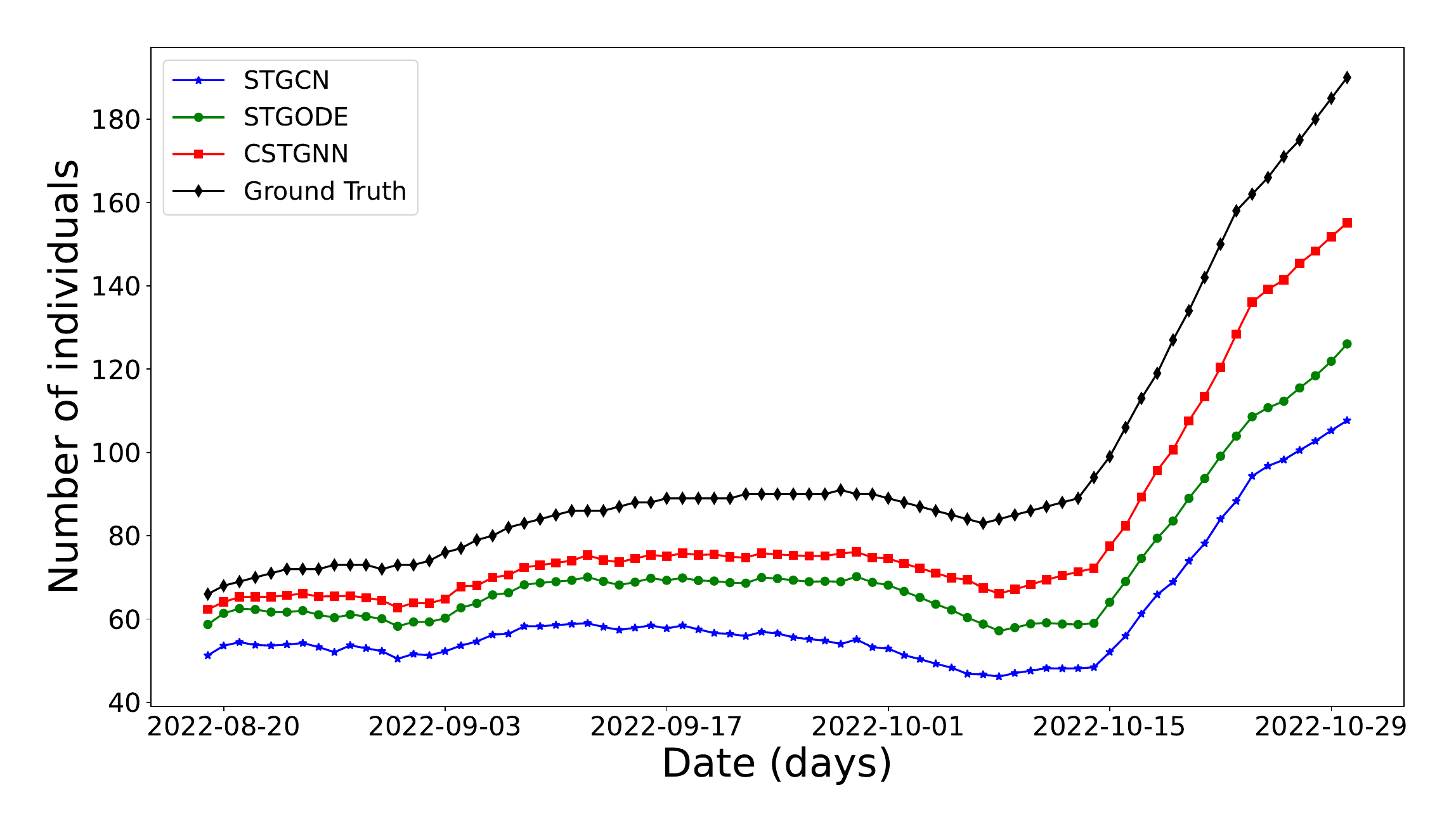}
        \captionsetup{font=normalsize}
        \caption*{(a) Beijingshi}
    \end{minipage}
    \hspace{-0.5cm} 
    \begin{minipage}[b]{0.498\textwidth}
        \centering
        \includegraphics[width=\textwidth]{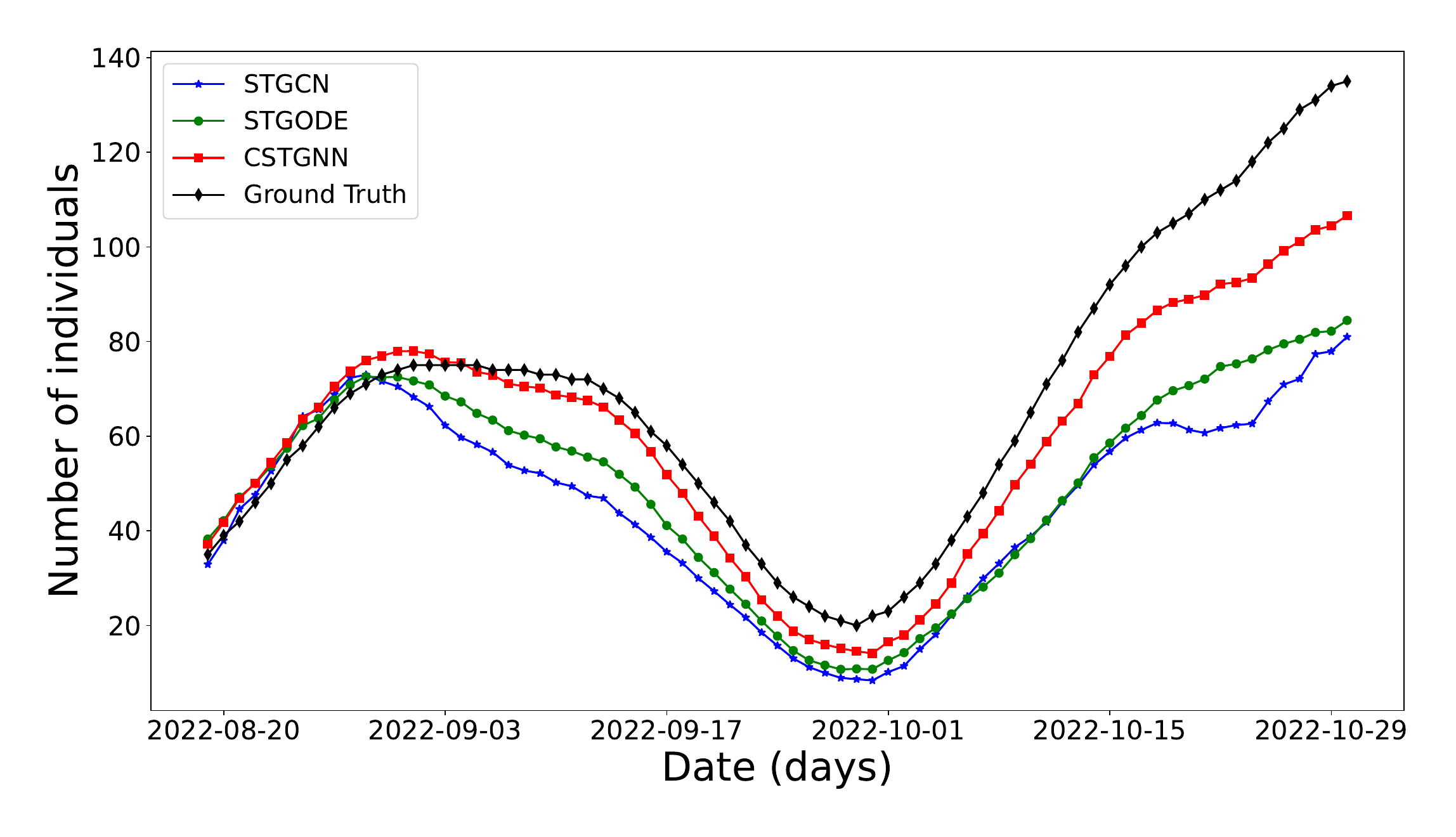}
        \captionsetup{font=normalsize}
        \caption*{(b) Chongqingshi}
    \end{minipage}
    \begin{minipage}[b]{0.498\textwidth}
        \centering
        \includegraphics[width=\textwidth]{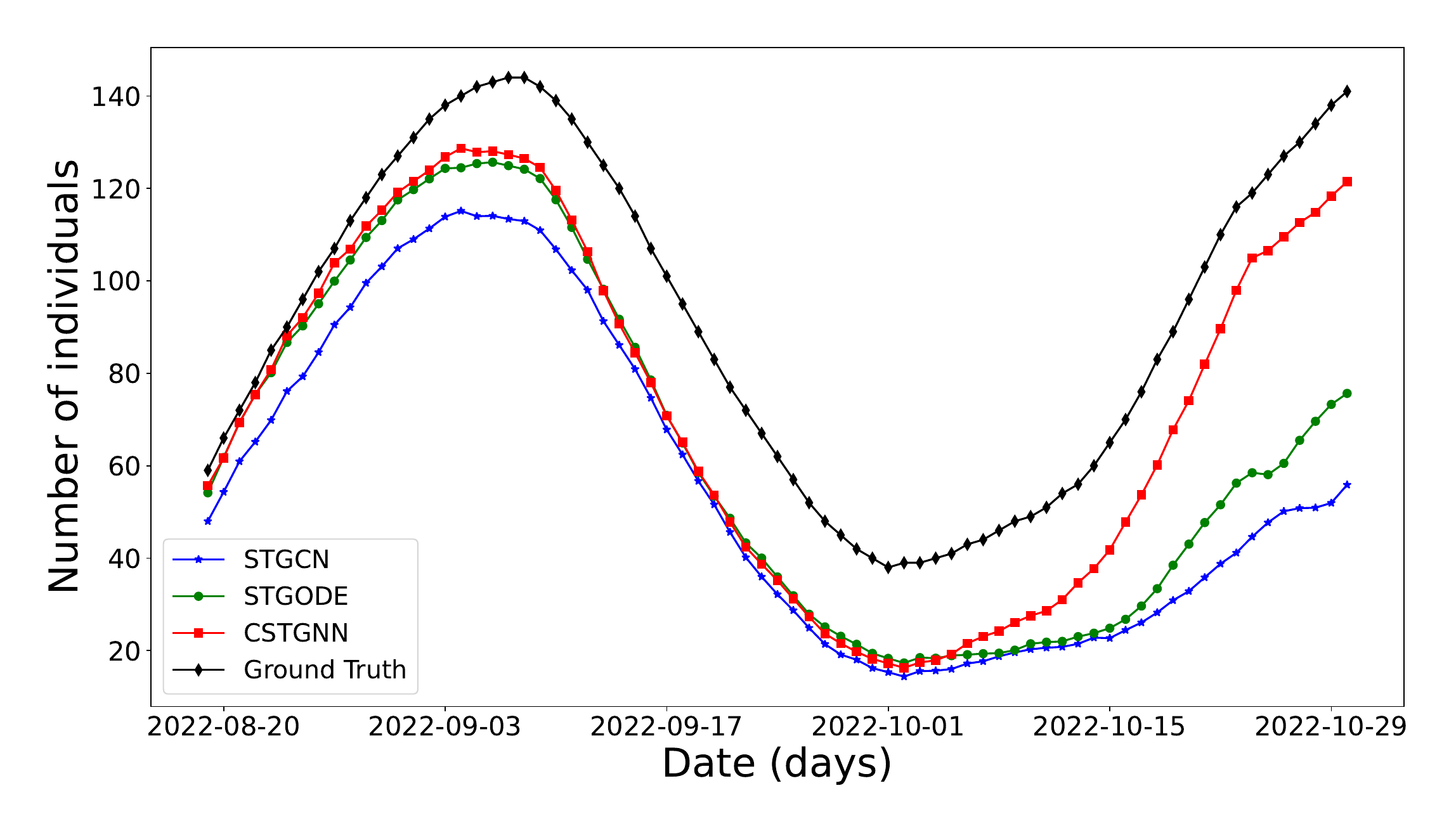}
        \captionsetup{font=normalsize}
        \caption*{(c) Shaanxisheng} 
    \end{minipage}
    \hspace{-0.5cm} 
    \begin{minipage}[b]{0.498\textwidth}
        \centering
        \includegraphics[width=\textwidth]{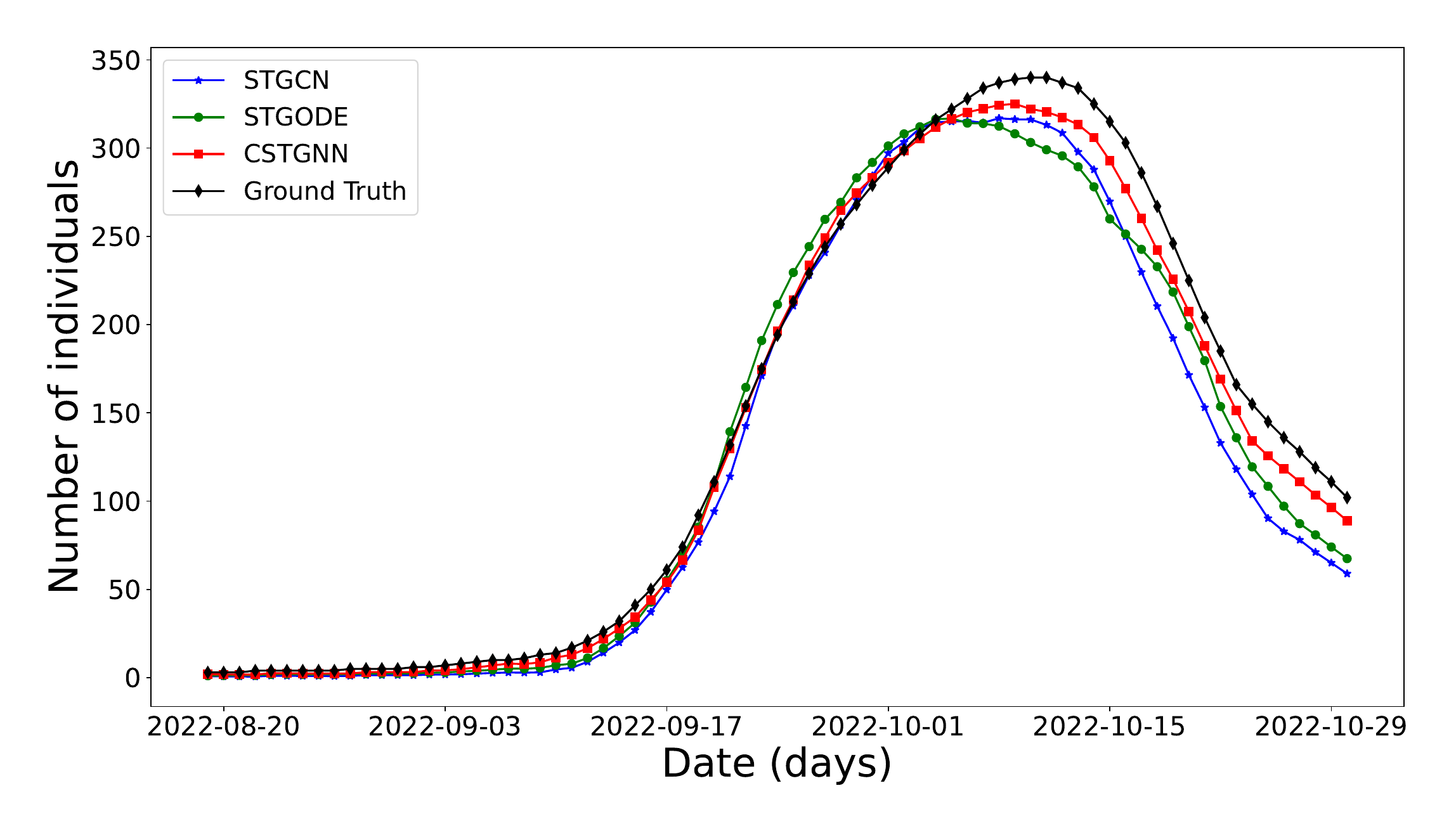}
        \captionsetup{font=normalsize}
        \caption*{(d) Guizhousheng}
    \end{minipage}
    \captionsetup{font=normalsize}
    \caption{Visualization of forecasting active case curves in representative provinces of China.}
    \label{fig:cc01}
\end{figure}

\begin{figure}[ht!]
    \centering
    \begin{minipage}[b]{0.498\textwidth} %
        \centering
        \includegraphics[width=\textwidth]{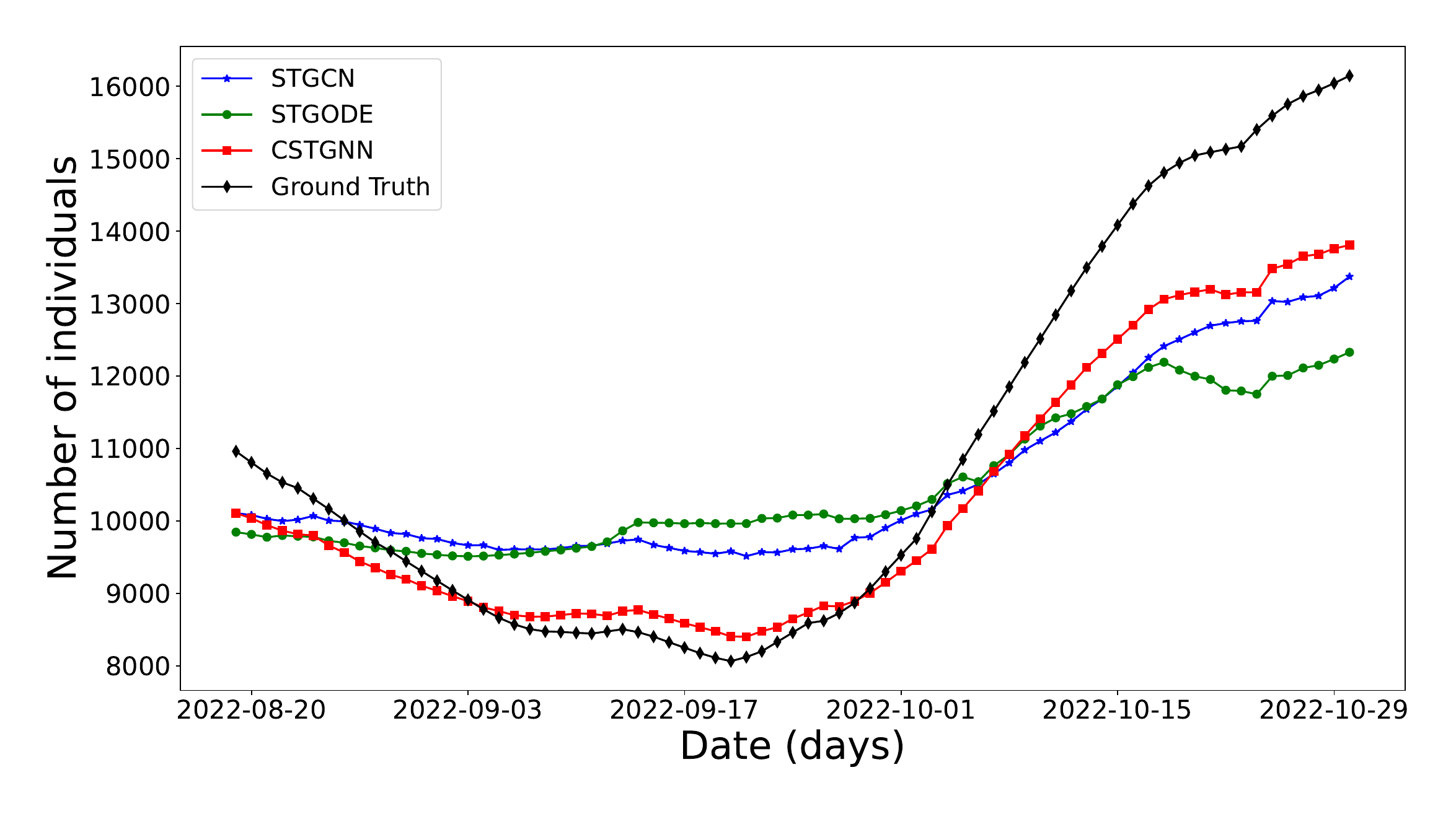}
        \captionsetup{font=normalsize}
        \caption*{(a) Berlin}
    \end{minipage}
    \hspace{-0.5cm} 
    \begin{minipage}[b]{0.498\textwidth} %
        \centering
        \includegraphics[width=\textwidth]{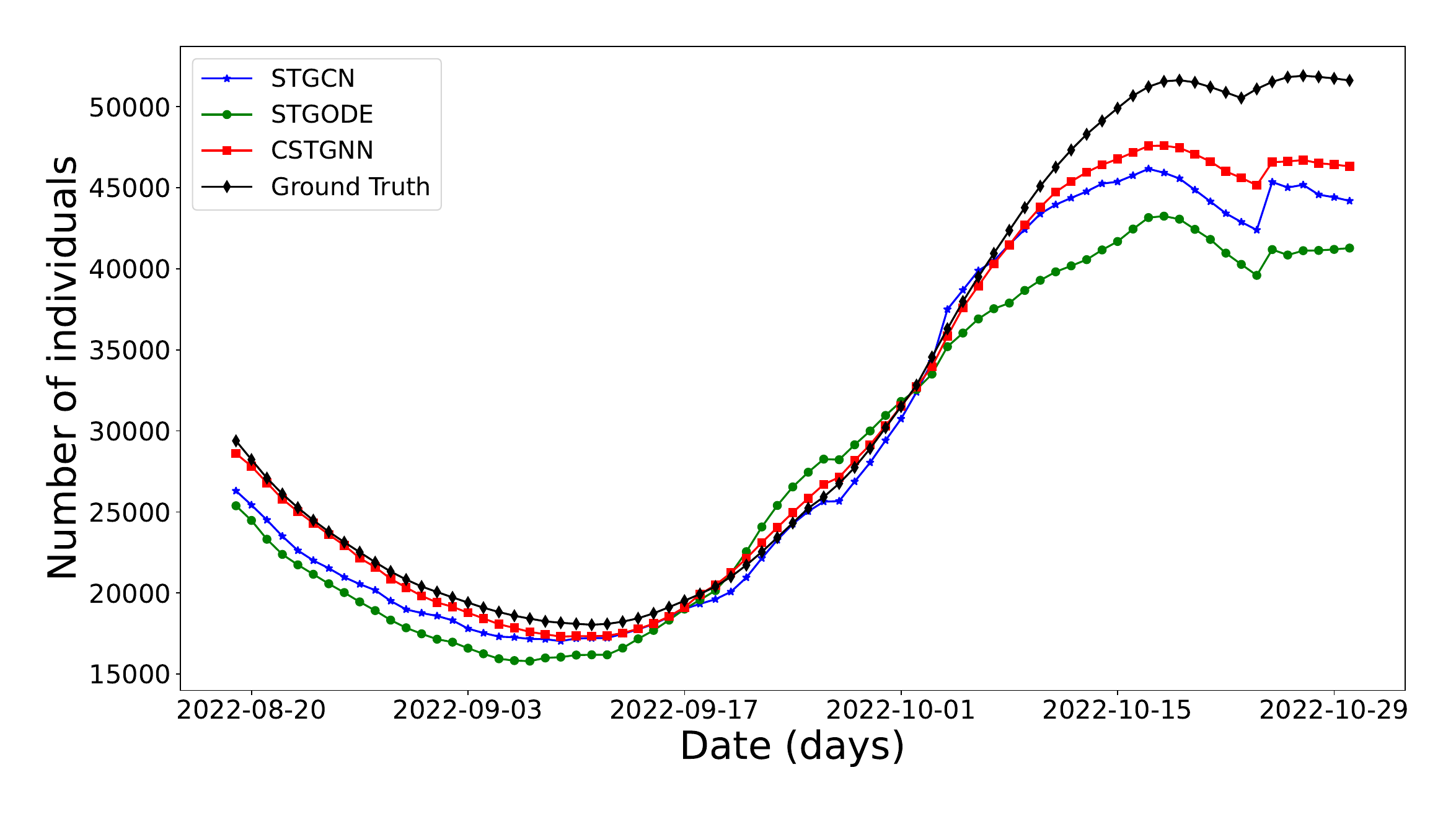}
        \captionsetup{font=normalsize}
        \caption*{(b) Hessen}
    \end{minipage}
    \begin{minipage}[b]{0.498\textwidth} %
        \centering
        \includegraphics[width=\textwidth]{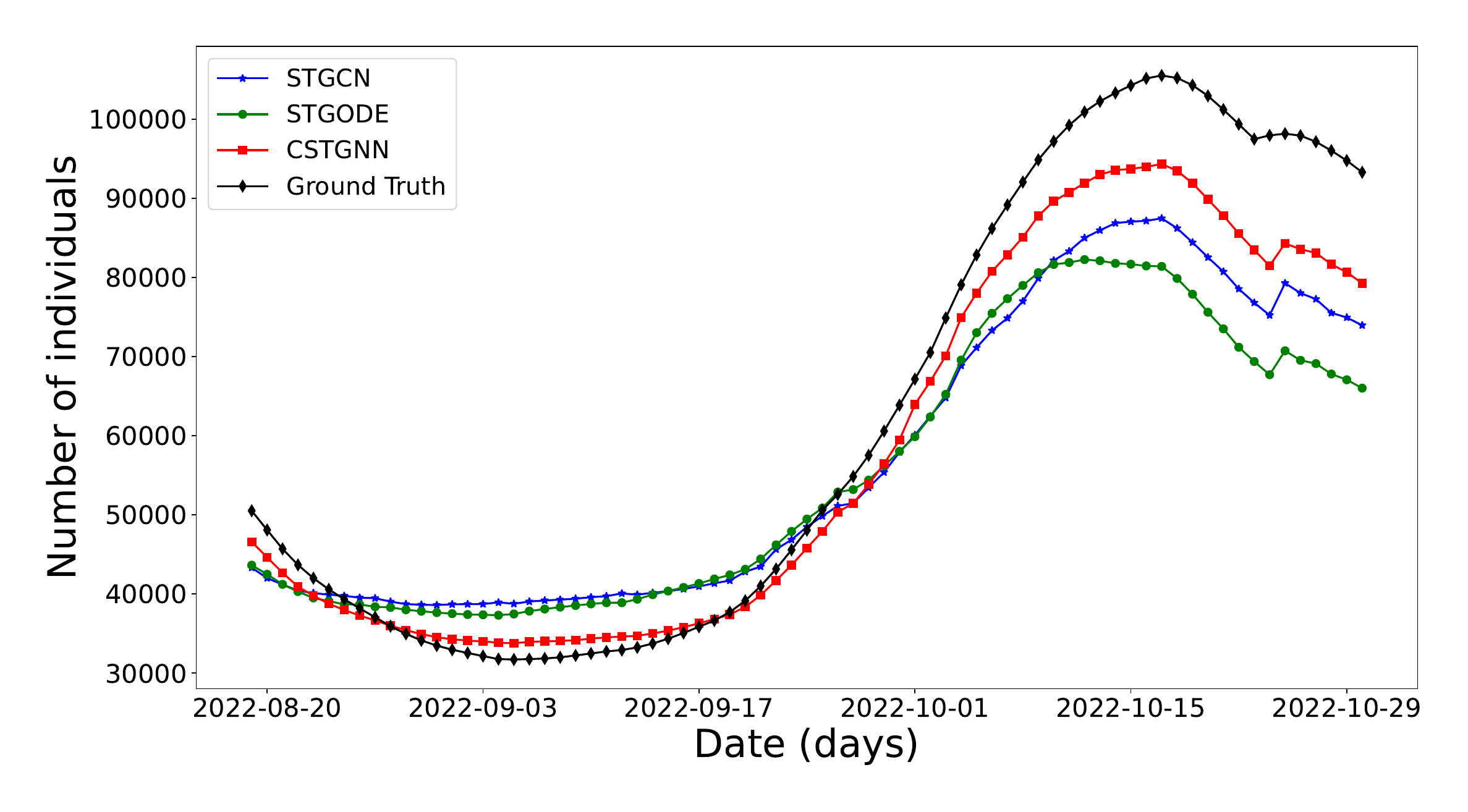}
        \captionsetup{font=normalsize}
        \caption*{(c) Bayern}
    \end{minipage}
    \hspace{-0.5cm} 
    \begin{minipage}[b]{0.498\textwidth} %
        \centering
        \includegraphics[width=\textwidth]{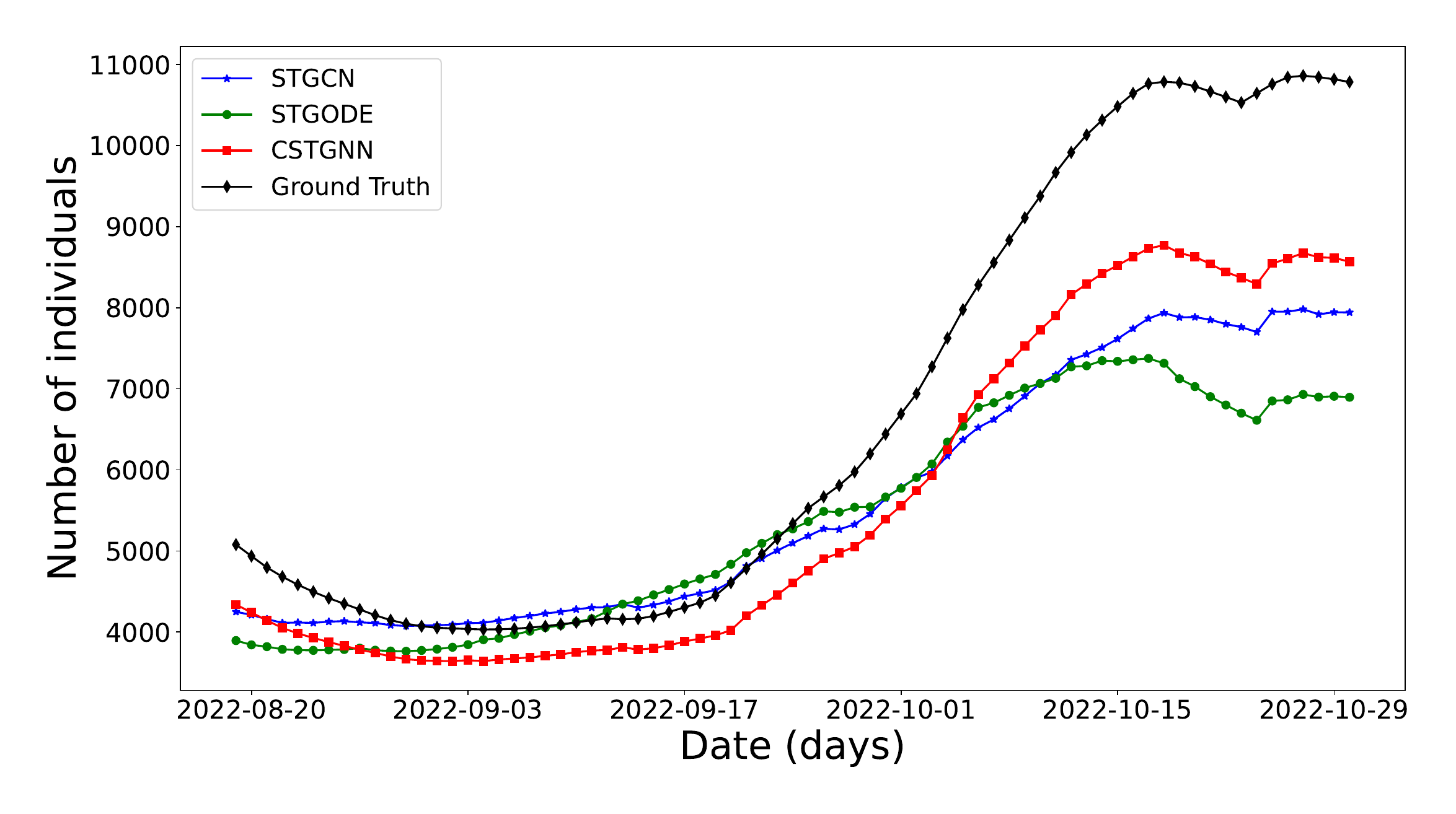}
        \captionsetup{font=normalsize}
        \caption*{(d) Thüringen}
    \end{minipage}
    \captionsetup{font=normalsize}
    \caption{Visualization of forecasting active case curves in representative states of Germany.}
    \label{fig:cg01}
\end{figure}

Moreover, selecting four representative provinces from the China dataset and four representative states from the Germany dataset clearly illustrates the comparison between the forecasted active cases and the reported values for a 7-day ahead prediction (L=7), with an emphasis on high-performing models. Figure~\ref{fig:cc01} displays the forecasting curves for Beijingshi (2022-08-19 to 2022-10-30), Chongqingshi, Shaanxisheng, and Guizhousheng in China. Figure~\ref{fig:cg01} shows the forecasting curves for Berlin (2022-08-19 to 2022-10-30), Hessen, Bayern, and Thüringen in Germany, respectively. Observe that the CSTGNN model’s forecasting curves are generally smoother and closer to the ground truth, across both datasets. For instance, the CSTGNN model accurately fits the upward trend of active cases, nearly overlapping with the reported values during the period from 2022-09-15 to 2022-09-30 in Guizhousheng and Hessen. Furthermore, as time progresses, the CSTGNN model shows superior performance in capturing long-term trends, in particular in the later stages (e.g., 2022-10-10 to 2022-10-30). Take, for example, Shaanxisheng and Thüringen: here, the CSTGNN model successfully captures the upwards or stabilizing trends in active cases during that period, whereas other models exhibit rather large deviations. \par 

These results demonstrate that the CSTGNN model does not only capture the dynamic trends efficiently and effectively in short-term forecasting but also excels in long-term predictions, by showing higher robustness and better accuracy by adapting to the evolving trends. Other models, in contrast, tend to diverge significantly when the reported values exhibit abrupt changes, e.g. the declining phase around 2022-10-01 in Chongqingshi. This observation is consistent with the results in Tables \ref{table:tbpe4china} and \ref{table:tbpe4germany}, further confirming the CSTGNN model’s superiority in forecasting tasks, particularly its ability to predict trends accurately as the forecasting horizon increases. \par 

The performance of different models is directly compared: The SIR model builds independent models for each region. It ignores the inter-regional interactions, which significantly limits its ability to capture the spatial dependence, which is so essential for accurate epidemic forecasting. The SCSIR model introduces time-varying contact parameters to account for inter-regional interactions. However, it still relies on predefined assumptions and simplified model formulations, which are insufficient to learn the complex and dynamic relationships between regions. In contrast, spatiotemporal deep learning models, such as CSTGNN, STGODE, and STGCN, learn dynamically both, inter-regional interactions and temporal patterns directly from the data. This enables these three models to adapt to evolving epidemic trends efficiently, thus resulting in significantly better performance in both, short-term and long-term forecasting. This is demonstrated for the China and Germany datasets (Tables \ref{table:tbpe4china} and \ref{table:tbpe4germany}). \par

Among deep learning-based models, while the STGCN model performs well on both datasets, its performance still lags behind that of the STGODE and CSTGNN models. This disparity may stem from the STGCN model’s limited consideration of epidemic transmission dynamics and domain knowledge. In contrast, spatiotemporal models that integrate the domain knowledge, such as STGODE and CSTGNN, demonstrate superior performance. Here, CSTGNN achieves the best results. The exceptional performance of CSTGNN can be attributed not only to its robust graph learning capabilities, but also to the incorporation of domain knowledge constraints. The contributions of graph learning and causal inference will be further discussed in Sections \ref{sec:eci} and \ref{sec:egl}. \par

\subsection{Effects of Causal Inference}
\label{sec:eci}
To investigate the impact of the epidemiological component on spatiotemporal epidemic forecasting, we designed two variant models: CSTGNN-CE (CSTGNN-Causal-Enabled) and CSTGNN-CF (CSTGNN-Causal-Free). These variants allow to evaluate the contribution of the causal inference mechanisms integrated into the CSTGNN model by introducing the following differences:\par

\begin{enumerate}[label=(\arabic*)]
\item \textbf{CSTGNN-CE}: The CSTGNN-CE model retains the causal inference module, which integrates domain knowledge of epidemic dynamics to constrain and guide the graph learning and forecasting process. This variant leverages causal relationships, to enhance spatiotemporal predictions, by modeling both inter-regional interactions and temporal patterns, influenced by epidemiological parameters.

\item \textbf{CSTGNN-CF}: The CSTGNN-CF model removes the causal inference module, excluding the epidemiological constraints and focusing solely on data-driven learning. By eliminating the causal components, this variant relies entirely on spatiotemporal feature extraction through graph learning and the temporal modules, without leveraging domain-specific epidemiological knowledge.
\end{enumerate}
This experimental setup facilitates the isolation and quantification of the impact of causal inference in the CSTGNN model, offering valuable insights into the role of domain knowledge in enhancing epidemic forecasting accuracy. \par

\begin{table}[ht!]
\centering
\captionsetup{font=normalsize}
\caption{Effects of Causal Inference on the China dataset.}
\label{table:tbeci4china}
\resizebox{\textwidth}{!}{
\begin{tabular}{@{}lllllllllll@{}}
\toprule
\multirow{3}{*}{} & \multicolumn{10}{l}{The China dataset} \\ \cmidrule(lr{0.1pt}){2-11} 
                  & \multicolumn{5}{l}{L=7} & \multicolumn{5}{l}{L=14} \\ 
                    \cmidrule(lr{0.1pt}){2-6} \cmidrule(lr{0.1pt}){7-11} 
Model                & MAE & RMSE & RAE & PCC & CCC & MAE & RMSE & RAE & PCC & CCC \\ \midrule
CSTGNN-CF            &44.3±13.2 &100.9±39.3 &0.3±0.1 &97.2±3.0\% &96.7±3.1\%
                     &71.47±13.72 &165.6±36.1 &0.4±0.1 &93.3±3.4\% &92.6±3.4\% \\ 
CSTGNN-CE            &39.6±5.3 &83.5±12.1 &0.2±0.0 &98.3±0.5\% &98.1±0.5\%  
                     &58.5±10.3 &147.7±27.2 &0.3±0.1 &95.4±1.3\% &93.8±3.6\% \\  \midrule
\multirow{2}{*}{} & \multicolumn{5}{l}{L=21} & \multicolumn{5}{l}{L=28} \\ 
                    \cmidrule(lr{0.1pt}){2-6} \cmidrule(lr{0.1pt}){7-11} 
Model                & MAE & RMSE & RAE & PCC & CCC & MAE & RMSE & RAE & PCC & CCC \\ \midrule
CSTGNN-CF            &98.6±21.9 &228.6±49.6 &0.6±0.1 &86.0±6.0\% &85.0±5.4\%
                     &127.6±43.2 &322.4±146.8 &0.7±0.3 &75.8±14.3\% &72.3±15.9\%  \\ 
CSTGNN-CE            &91.0±16.6 &213.1±31.8 &0.5±0.1 &88.3±4.7\% &84.70±7.7\%
                     &111.7±44.1 &270.9±100.9 &0.6±0.3 &75.0±23.8\% &73.5±23.0\% \\ \midrule
\end{tabular}
}
\end{table}

The experimental results on the China and Germany datasets are shown in Table \ref{table:tbeci4china} and Table \ref{table:tbeci4germany}, respectively. On the China dataset, CSTGNN-CE consistently outperforms CSTGNN-CF across all forecasting horizons. By incorporating causal inference, CSTGNN-CE leverages domain-specific epidemiological knowledge to capture complex spatio-temporal dependence more effectively. For instance, at L = 7 and L = 14, CSTGNN-CE achieves notable improvements in MAE and RMSE, while also showing higher PCC and CCC values, demonstrating stronger alignment with the actual epidemic trends. These results emphasize the advantage of integrating causal inference in achieving more accurate and reliable epidemic forecasts.\par

\begin{table}[ht!]
\centering
\captionsetup{font=normalsize}
\caption{Effects of Causal Inference on the Germany dataset.}
\label{table:tbeci4germany}
\resizebox{\textwidth}{!}{
\begin{tabular}{@{}lllllllllll@{}}
\toprule
\multirow{3}{*}{} & \multicolumn{10}{l}{The Germany dataset} \\ \cmidrule(lr{0.1pt}){2-11} 
                  & \multicolumn{5}{l}{L=7} & \multicolumn{5}{l}{L=14} \\ 
                    \cmidrule(lr{0.1pt}){2-6} \cmidrule(lr{0.1pt}){7-11} 
Model                & MAE & RMSE & RAE & PCC & CCC & MAE & RMSE & RAE & PCC & CCC \\ \midrule
CSTGNN-CF            &7447.2±0.00 &13069.1±0.00 &0.4±0.0 &90.85±0.0\% &82.2±0.0\%
                     &8951.3±0.00 &159237±0.00 &0.5±0.0 &84.55±0.0\% &70.8±0.0\% \\ 
CSTGNN-CE            &4393.8±416.8 &7511.6±184.2 &0.2±0.1 &97.66±1.3\% &94.4±2.5\%
                     &7400.7±949.6 &12577.1±361.6 &0.4±0.1 &92.66±2.0\% &80.8±11.1\% \\  \midrule
\multirow{2}{*}{} & \multicolumn{5}{l}{L=21} & \multicolumn{5}{l}{L=28} \\ 
                    \cmidrule(lr{0.1pt}){2-6} \cmidrule(lr{0.1pt}){7-11} 
Model                & MAE & RMSE & RAE & PCC & CCC & MAE & RMSE & RAE & PCC & CCC \\ \midrule
CSTGNN-CF            &8699.4±664.9 &14509.7±363.5 &0.5±0.1 &87.7±4.0\% &79.5±6.5\%
                     &8914.0±987.4 &15017.5±765.7 &0.6±0.1 &85.6±4.8\% &74.3±11.8\% \\
CSTGNN-CE            &7487.5±425.9 &12220.5±931.9 &0.4±0.1 &89.0±3.4\% &86.0±4.1\%
                     &8423.9±422.2 &13880.9±1406.2 &0.5±0.1 &86.2±3.7\% &77.0±15.7\% \\  \midrule
\end{tabular}
}
\end{table}

Similarly, on the German dataset, CSTGNN-CE demonstrates outstanding performance in both short-term and long-term prediction tasks. For instance, at L = 7, L = 14, and L = 21, CSTGNN-CE achieves substantial reductions in MAE and RMSE compared to CSTGNN-CF, highlighting its robustness in handling extended prediction horizons. In contrast, while CSTGNN-CF performs relatively better at L = 7 compared to other prediction windows, its reliance on purely data-driven learning limits its ability to effectively capture long-term dependence. This limitation results in a pronounced decline in performance as the prediction window expands. Overall, the integration of causal inference empowers CSTGNN-CE to deliver more stable and accurate forecasts, consistently surpassing CSTGNN-CF across diverse datasets and prediction scenarios. \par

\subsection{Effects of Graph Learning}
\label{sec:egl}
To investigate the impact of the human mobility component on spatio-temporal epidemic forecasting, we again devised two variant models: CSTGNN-Dynamic (CSTGNN with Dynamic Graph Learning) and CSTGNN-Static (CSTGNN with Static Adjacency Matrix). These variants allow us to evaluate the effectiveness of simulating population mobility patterns through dynamic graph learning compared to using a fixed adjacency matrix, and differ as follows: \par

\begin{enumerate}[label=(\arabic*)]
\item \textbf{CSTGNN-Dynamic}: This variant incorporates a dynamic graph learning mechanism to simulate population mobility patterns as a time-varying adjacency matrix. The dynamic adjacency matrix is learned through the graph learning algorithm and represents the evolving interactions between regions over time. By modeling these dynamic changes, this approach aims to approximate real-world population movements, allowing the GCN to process accurate and temporally adaptive spatial relationships.

\item \textbf{CSTGNN-Static}: In this variant, the graph structure is represented by a static binary adjacency matrix, where edges are pre-defined based on geographical proximity (1 for neighboring regions and 0 otherwise). While this approach provides a straightforward and computationally efficient alternative, it assumes fixed spatial relationships and does not account for temporal variations in population mobility, potentially limiting its ability to adapt to dynamic epidemic patterns.
\end{enumerate}
This experimental setup facilitates the isolation and quantification of the impact of dynamic graph learning in CSTGNN, providing insights into the role of simulated temporally adaptive spatial relationships in enhancing epidemic forecasting accuracy. \par

\begin{table}[ht!]
\centering
\captionsetup{font=normalsize}
\caption{Effects of Graph Learning on the China dataset.}
\label{table:tbegl4china}
\resizebox{\textwidth}{!}{
\begin{tabular}{@{}lllllllllll@{}}
\toprule
\multirow{3}{*}{} & \multicolumn{10}{l}{The China dataset} \\ \cmidrule(lr{0.1pt}){2-11} 
                  & \multicolumn{5}{l}{L=7} & \multicolumn{5}{l}{L=14} \\ 
                    \cmidrule(lr{0.1pt}){2-6} \cmidrule(lr{0.1pt}){7-11} 
Model                & MAE & RMSE & RAE & PCC & CCC & MAE & RMSE & RAE & PCC & CCC \\ \midrule
CSTGNN-Static        &44.4±11.3 &100.0±32.2 &0.3±0.1 &97.2±2.4\% &96.9±2.4\%
                     &64.8±8.9 &154.2±26.1 &0.4±0.1 &94.4±1.8\% &93.8±1.9\% \\ 
CSTGNN-Dynamic       &39.6±5.3 &83.5±12.1 &0.2±0.0 &98.3±0.5\% &98.1±0.5\%  
                     &58.5±10.3 &147.7±27.2 &0.3±0.1 &95.4±1.3\% &93.8±3.6\% \\  \midrule
\multirow{2}{*}{} & \multicolumn{5}{l}{L=21} & \multicolumn{5}{l}{L=28} \\ 
                    \cmidrule(lr{0.1pt}){2-6} \cmidrule(lr{0.1pt}){7-11} 
Model                & MAE & RMSE & RAE & PCC & CCC & MAE & RMSE & RAE & PCC & CCC \\ \midrule
CSTGNN-Static        &116.0±42.7 &279.4±122.4 &0.7±0.2 &81.3±10.8\% &78.8±12.3\%
                     &116.0±31.6 &276.8±80.4 &0.7±0.2 &79.4±8.8\% &78.3±9.8\% \\ 
CSTGNN-Dynamic       &91.0±16.6 &213.1±31.8 &0.5±0.1 &88.3±4.7\% &84.7±7.7\%
                     &111.7±44.1 &270.9±100.9 &0.6±0.3 &75.0±23.8\% &73.5±23.0\% \\ \midrule
\end{tabular}
}
\end{table}

A comparison of our experimental results for both models operating on the China and Germany datasets, shown in Table \ref{table:tbegl4china} and Table \ref{table:tbegl4germany}, highlights the impact of dynamic graph learning mechanisms on spatio-temporal epidemic forecasting. On both datasets, the CSTGNN-Dynamic model, which utilizes a graph learning module designed to learn population mobility patterns as a dynamic adjacency matrix, consistently outperforms the CSTGNN-Static model. For example, for the China dataset, at L = 14, CSTGNN-Dynamic achieves an MAE of 58.5 compared to 64.8 for CSTGNN-Static, along with notable improvements in RMSE, PCC, and CCC, demonstrating its ability to effectively model dynamic spatial relationships. Similarly, for the Germany dataset, at L = 28, CSTGNN-Dynamic achieves a CCC of 77\%, significantly higher than the 71.1\% achieved by CSTGNN-Static, further validating the benefits of learning and simulating population mobility patterns to construct adaptive adjacency matrices. \par

\begin{table}[ht!]
\centering
\captionsetup{font=normalsize}
\caption{Effects of Graph Learning on the Germany dataset.}
\label{table:tbegl4germany}
\resizebox{\textwidth}{!}{
\begin{tabular}{@{}lllllllllll@{}}
\toprule
\multirow{3}{*}{} & \multicolumn{10}{l}{The Germany dataset} \\ \cmidrule(lr{0.1pt}){2-11} 
                  & \multicolumn{5}{l}{L=7} & \multicolumn{5}{l}{L=14} \\ 
                    \cmidrule(lr{0.1pt}){2-6} \cmidrule(lr{0.1pt}){7-11} 
Model                & MAE & RMSE & RAE & PCC & CCC & MAE & RMSE & RAE & PCC & CCC \\ \midrule
CSTGNN-Static        &5709.5±2841.3 &9294.1±4755.8 &0.3±0.2 &95.5±3.8\% &88.9±11.8\%
                     &7480.5±2640.8 &12565.5±4267.2 &0.4±0.2 &91.6±3.5\% &80.0±14.5\% \\ 
CSTGNN-Dynamic       &4393.8±416.8 &7511.6±184.2 &0.2±0.1 &97.7±1.3\% &94.4±2.5\%
                     &7400.7±949.6 &12577.1±361.6 &0.4±0.1 &92.7±1.9\% &80.8±11.1\% \\  \midrule
\multirow{2}{*}{} & \multicolumn{5}{l}{L=21} & \multicolumn{5}{l}{L=28} \\ 
                    \cmidrule(lr{0.1pt}){2-6} \cmidrule(lr{0.1pt}){7-11} 
Model                & MAE & RMSE & RAE & PCC & CCC & MAE & RMSE & RAE & PCC & CCC \\ \midrule
CSTGNN-Static        &8729.3±2594.2 &14413.1±4162.4 &0.5±0.2 &85.4±4.8\% &81.1±7.0\%
                     &9230.7±3171.6 &15435.2±5000.7 &0.5±0.2 &85.6±3.4\% &71.1±22.1\% \\ 
CSTGNN-Dynamic       &7487.5±425.9 &12220.5±931.9 &0.4±0.1 &89.0±3.4\% &86.0±4.1\%
                     &8423.9±422.2 &13880.9±1406.2 &0.5±0.1 &86.2±3.7\% &77.0±15.7\% \\  \midrule
\end{tabular}
}
\end{table}
In contrast, the CSTGNN-Static model, which uses a static binary adjacency matrix based on geographical proximity, performs reasonably well but lacks the flexibility to adapt to temporal variations in spatial relationships. This limitation becomes more apparent in longer forecasting horizons, where the ability to model dynamic and evolving interactions between regions is critical. By incorporating a graph learning mechanism to simulate population mobility patterns as a dynamic adjacency matrix, CSTGNN-Dynamic demonstrates greater adaptability and accuracy. Overall, the use of dynamic graph learning enables CSTGNN-Dynamic to deliver more robust and reliable predictions, highlighting the importance of adaptive graph structures in spatio-temporal epidemic forecasting. \par

\subsection{Case Study}
\label{sec:cs}
The Spatio-Contact SIR model, as an essential part of our framework, significantly enhances the interpretability of the results generated by our framework. By analyzing the learned epidemic parameters ($\beta$, $\gamma$, and $c$) generated by the Spatio-Temporal module, we demonstrate how the model captures meaningful interaction patterns between regions, reflecting real-world population mobility and its spatio-temporal dynamics. We base our analyses on the 7-day ahead prediction results in the best-performing model, further demonstrating how the module enhances the framework’s interpretability through its ability to model spatio-temporal interactions.

\subsubsection{Contact Rate}
\begin{figure}[htbp!]
    \centering
    \begin{minipage}[b]{0.24\textwidth}
        \centering
        \includegraphics[width=\textwidth]{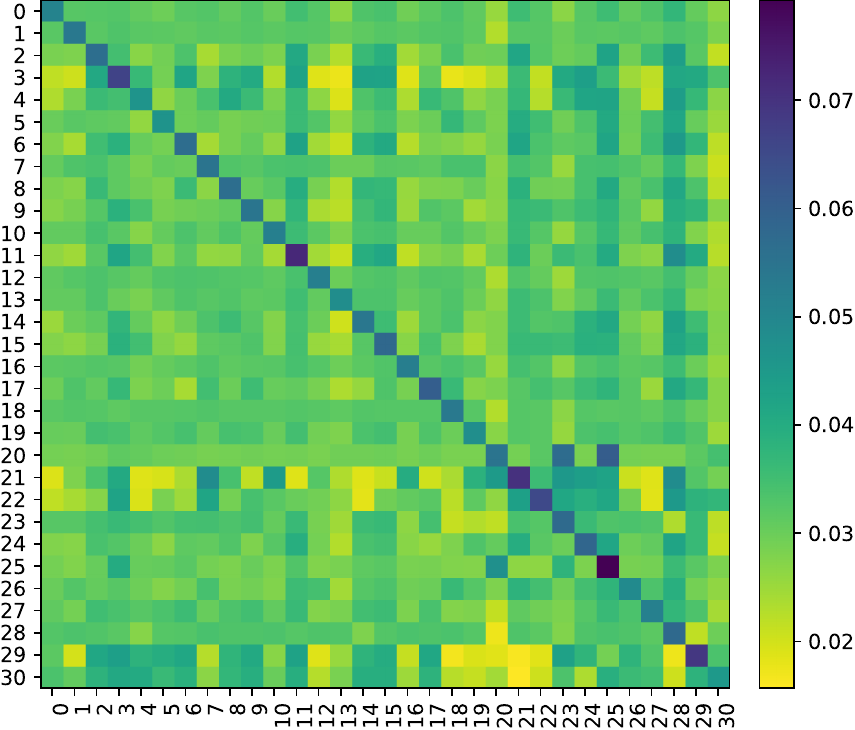}
        \captionsetup{font=normalsize}
        \caption*{(a) 2022-08-31}
    \end{minipage}
    \begin{minipage}[b]{0.24\textwidth}
        \centering
        \includegraphics[width=\textwidth]{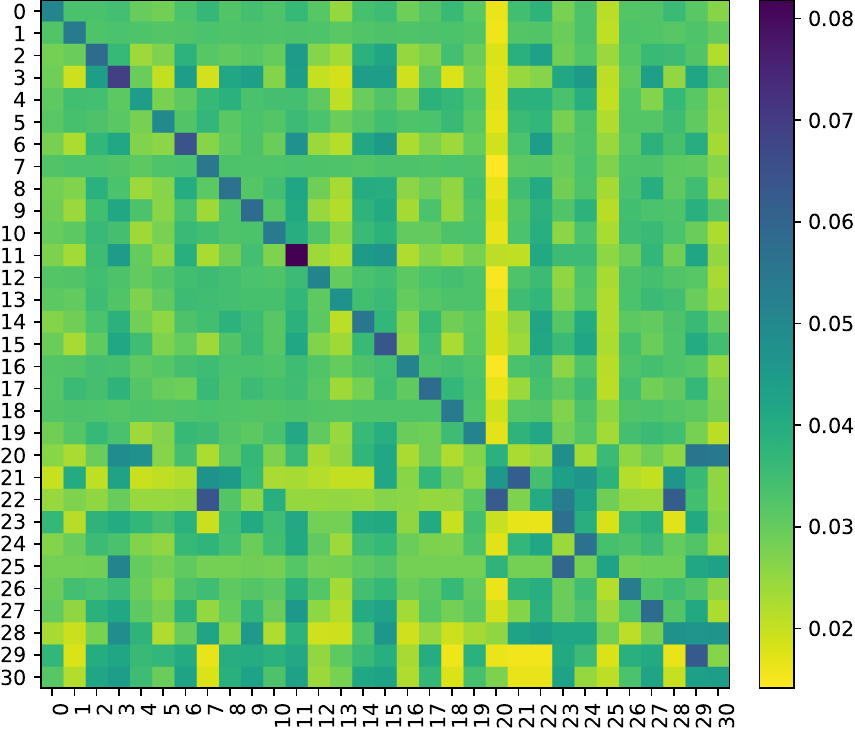}
        \captionsetup{font=normalsize}
        \caption*{(b) 2022-09-10}
    \end{minipage}
    \begin{minipage}[b]{0.24\textwidth}
        \centering
        \includegraphics[width=\textwidth]{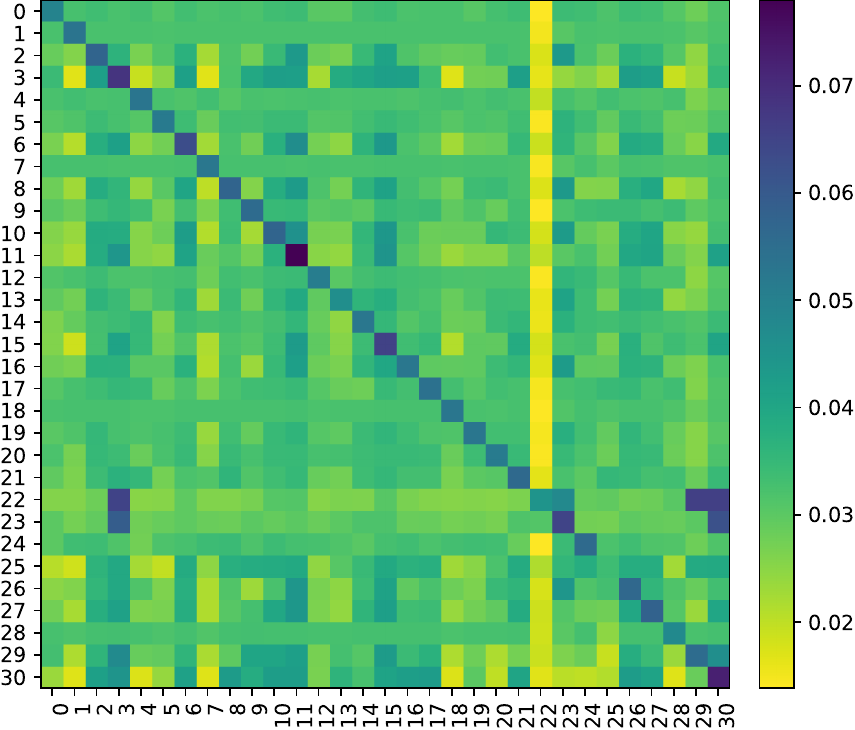}
        \captionsetup{font=normalsize}
        \caption*{(c) 2022-10-05}
    \end{minipage}
    \begin{minipage}[b]{0.24\textwidth}
        \centering
        \includegraphics[width=\textwidth]{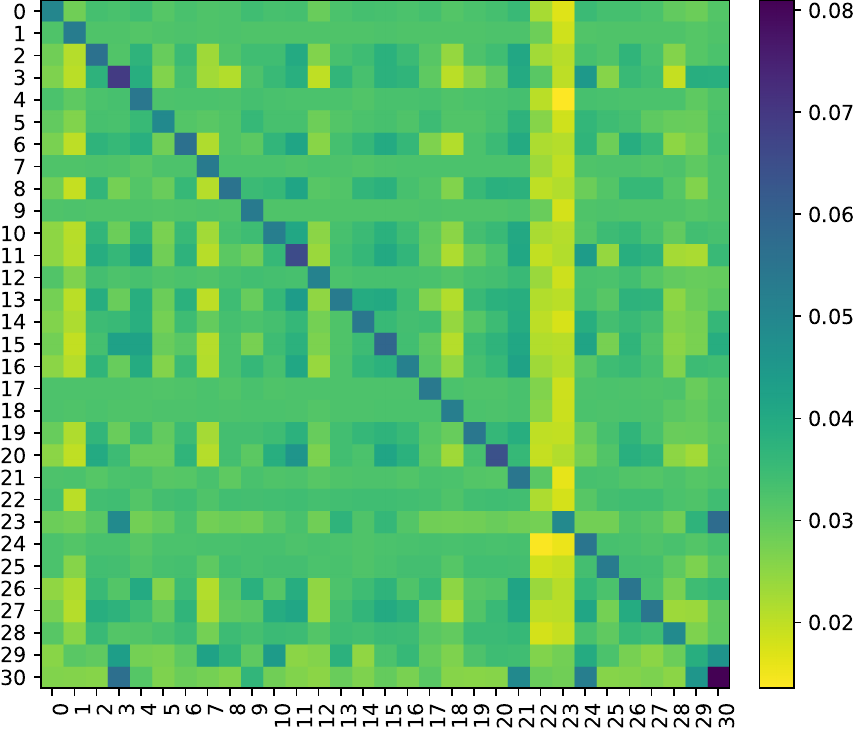}
        \captionsetup{font=normalsize}
        \caption*{(d) 2022-10-13}
    \end{minipage}
    \captionsetup{font=normalsize}
    \caption{Temporal Dynamics of Learned Contact Rates Across Provinces in China.}
    \label{fig:cslcrc}
\end{figure}

The contact rates in China exhibit distinct temporal and spatial variations influenced by holidays and regional interactions. As shown in Figure~\ref{fig:cslcrc}, intraregional interactions (diagonal elements) remain dominant across all dates, reflecting the prevalence of localized mobility. Off-diagonal elements, however, reveal notable interprovincial interactions during specific periods. On August 31, 2022, marking the end of summer vacations, elevated contact rates are observed between densely populated areas such as Beijingshi (index 0), Tianjinshi~(1), and Hebeisheng~(2), corresponding to student and family travel. Coastal regions like Zhejiangsheng~(10) and Fujian~(12) also demonstrate significant interactions, driven by their status as popular tourist destinations. During the \emph{Mid-Autumn Festival} on September 10, 2022, regions with strong familial or cultural ties, such as Guangdongsheng~(18) and Hainansheng~(20), show increased connectivity, highlighting family reunions. Figure~\ref{fig:cslcrc}(c) illustrates October 5, 2022, during the \emph{National Day Golden Week} when interprovincial mobility peaks, particularly among tourist hubs like Shanghaishi~(8), Zhejiangsheng~(10), and Guangdongsheng~(18). By October 13, 2022, as the holiday season concludes, interregional interactions decline slightly, with economic centers like Shanghaishi~(8) and Guangdongsheng~(18) sustaining higher contact rates due to commuting and economic activity. \par

\begin{figure}[htbp!]
    \centering
    \begin{minipage}[b]{0.24\textwidth}
        \centering
        \includegraphics[width=\textwidth]{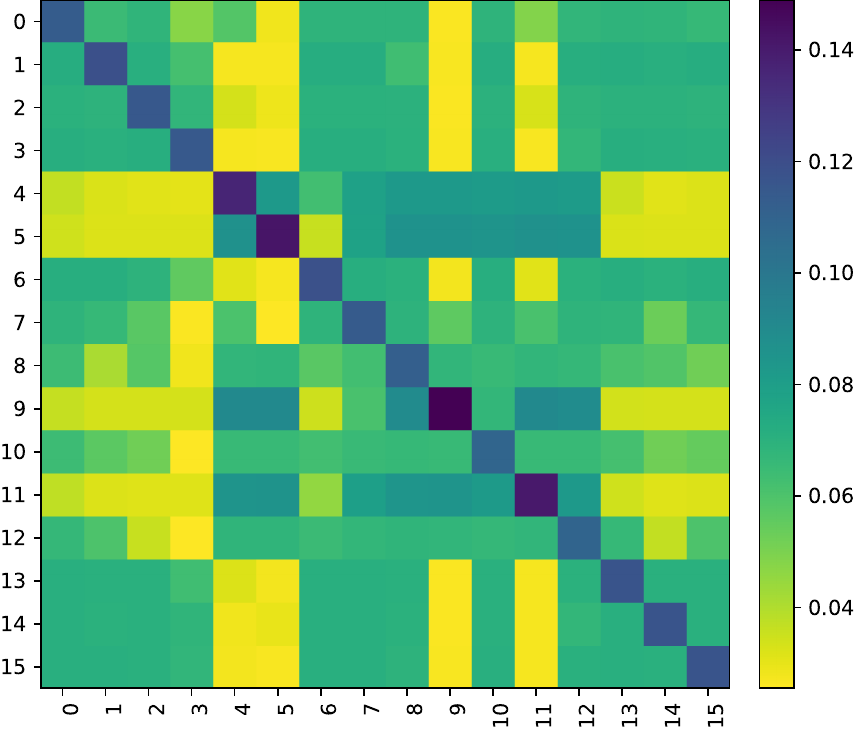}
        \captionsetup{font=normalsize}
        \caption*{(a) 2022-08-29}
    \end{minipage}
    \begin{minipage}[b]{0.24\textwidth}
        \centering
        \includegraphics[width=\textwidth]{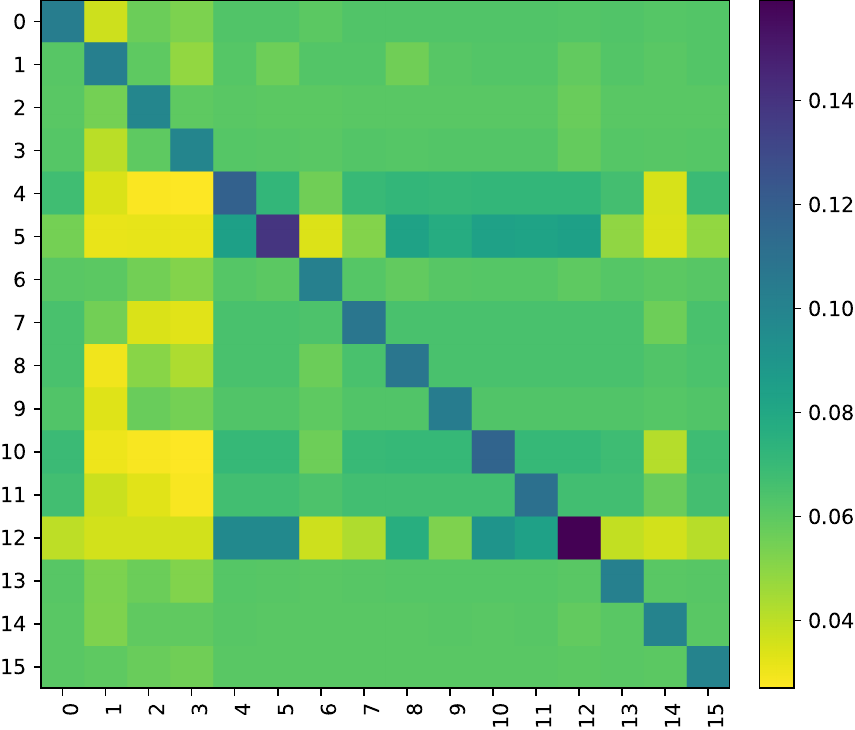}
        \captionsetup{font=normalsize}
        \caption*{(b) 2022-09-26}
    \end{minipage}
    \begin{minipage}[b]{0.24\textwidth}
        \centering
        \includegraphics[width=\textwidth]{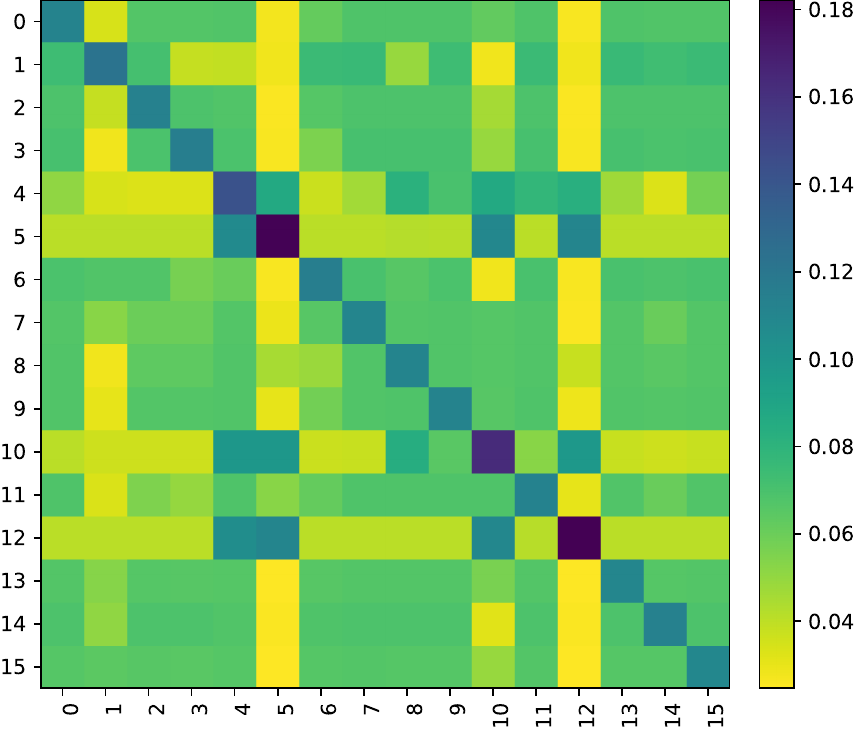}
        \captionsetup{font=normalsize}
        \caption*{(c) 2022-10-03}
    \end{minipage}
    \begin{minipage}[b]{0.24\textwidth}
        \centering
        \includegraphics[width=\textwidth]{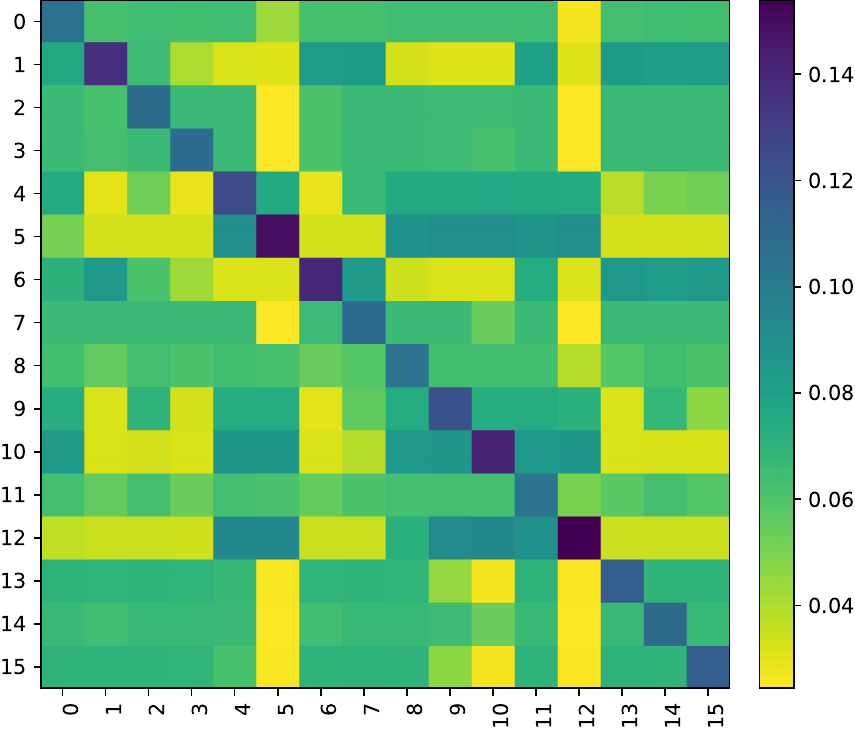}
        \captionsetup{font=normalsize}
        \caption*{(d) 2022-10-30}
    \end{minipage}
    \captionsetup{font=normalsize}
    \caption{Temporal Dynamics of Learned Contact Rates Across Federal States in Germany.}
    \label{fig:cslcrg}
\end{figure}

For Germany, the learned contact rates reflect the effects of holidays, cultural events, and regular commuting on mobility patterns. As shown in Figure~\ref{fig:cslcrg}, intraregional interactions (diagonal elements) dominate consistently, underscoring the importance of localized movements within states. On August 29, 2022, at the end of summer breaks, Berlin (index 1) and Brandenburg (0) show heightened interactions due to commuting and vacation-related travel. Similarly, coastal regions like Schleswig-Holstein~(11) exhibit higher contact rates, driven by their role as popular summer destinations. On September 26, 2022, during the \emph{Oktoberfest} in Bayern~(3), interregional interactions rise sharply, particularly between Bayern and neighboring states such as Baden-Württemberg~(2) and Hessen~(5), reflecting festival-driven population flows. Figure~\ref{fig:cslcrg}(b) highlights this spike, with darker off-diagonal elements indicating substantial interregional mobility. October 3, 2022, associated with Germany’s Unity Day, shows moderate interregional travel, with populous states like Nordrhein-Westfalen~(9) and Rheinland-Pfalz~(10) contributing notably to increased connectivity. By October 30, 2022, interactions stabilize, with economically significant states like Bayern~(3) and Nordrhein-Westfalen~(9) maintaining higher intraregional activity due to their economic roles and population density. \par

These findings confirm the critical role of population mobility, driven by holidays, cultural events, and economic activities, in shaping the dynamics of contact rates. The Spatio-Contact SIR model effectively captures these variations, demonstrating its ability to align learned parameters with real-world mobility patterns. By identifying both localized and interregional trends, the model enhances interpretability and provides valuable insights for epidemic modeling and public health policy planning. \par

\subsubsection{Effective Reproduction Number}
To enhance the interpretability of the proposed spatio-temporal SIR model, we devised a new definition of the effective reproduction number $R_0(t)$. Unlike the traditional basic reproduction number $R_0$, which assumes homogeneous mixing and static conditions, our $R_0(t)$ incorporates spatial heterogeneity and temporal dynamics. This is achieved by integrating learned parameters into a time-dependent next-generation matrix \cite{Diekmann1990}, providing a more flexible and context-sensitive measure of instantaneous transmission potential. In classical epidemic models, $R_0$ is given by:
\begin{equation}
    R_0 = \frac{\beta}{\gamma}
\end{equation}
where $\beta$ denotes the infection rate and $\gamma$ is the recovery rate. This ratio represents the initial growth potential of an outbreak in a fully susceptible population. However, such a definition does not account for regional interactions or time-varying conditions.

In our framework, transmission patterns vary dynamically across both space and time. To approximate a meaningful $R_0(t)$, we construct a time-dependent next-generation matrix that incorporates these variations. To this end, we first define a diagonal matrix at time $t$,
\begin{equation}
    D(t) = \operatorname{diag}\left(\frac{\beta_1(t)}{\gamma_1(t)}, \frac{\beta_2(t)}{\gamma_2(t)}, \dots, \frac{\beta_Q(t)}{\gamma_Q(t)}\right),
\end{equation}
where each element represents the transmission-to-recovery ratio for a specific region. We also define a contact matrix
\begin{equation}
C(t) = (c_{ij}(t)), \quad i, j = 1, 2, \dots, Q,
\label{eq:C_matrix}
\end{equation}
where $c_{ij}(t)$ denotes the contact intensity from region $j$ to region $i$. The next-generation matrix is then computed as
\begin{equation}
    M(t) = D(t)C(t)
\end{equation}
where $M_{ij}(t)$ represents the expected number of secondary infections in region $i$ caused by a single infected individual in region $j$, assuming a fully susceptible population. Following the next-generation matrix approach \cite{Diekmann1990}, the effective reproduction number $R_0(t)$ is then determined as the spectral radius (i.e., the largest eigenvalue) of $M(t)$:
\begin{equation}
    R_0(t) = \lambda_{\max}(M(t)).
\end{equation}
This represents the dominant transmission potential in a structured population. This formulation encapsulates both spatial heterogeneity and temporal dynamics, providing a robust and adaptive estimate of the instantaneous transmission potential. \par

\begin{figure}[ht!]
    \centering
    \includegraphics[width=17.5cm]{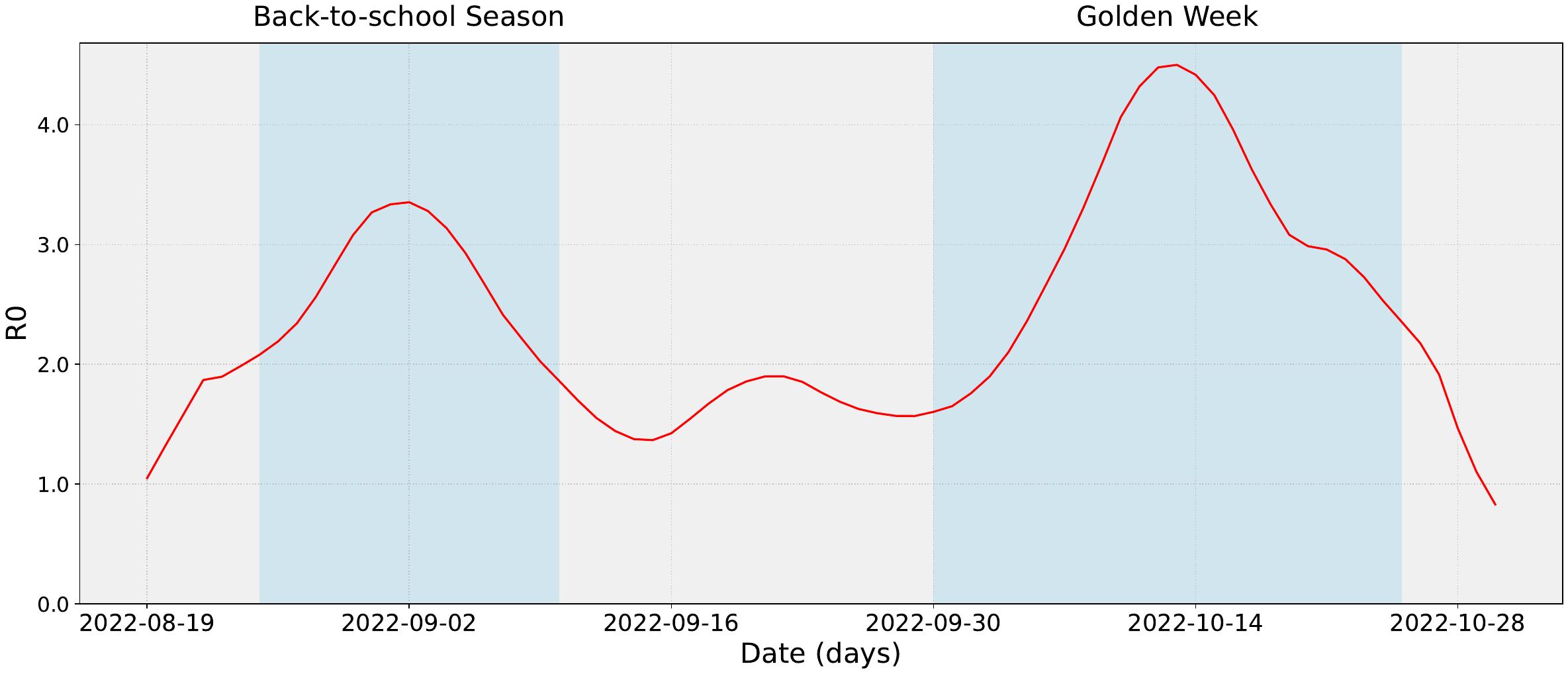}
    \captionsetup{font=normalsize}
    \caption{Visualization of $R_0$ curves in Shaanxisheng of China.}
    \label{fig:ernc}
\end{figure}

\begin{figure}[ht!]
    \centering
    \includegraphics[width=17.5cm]{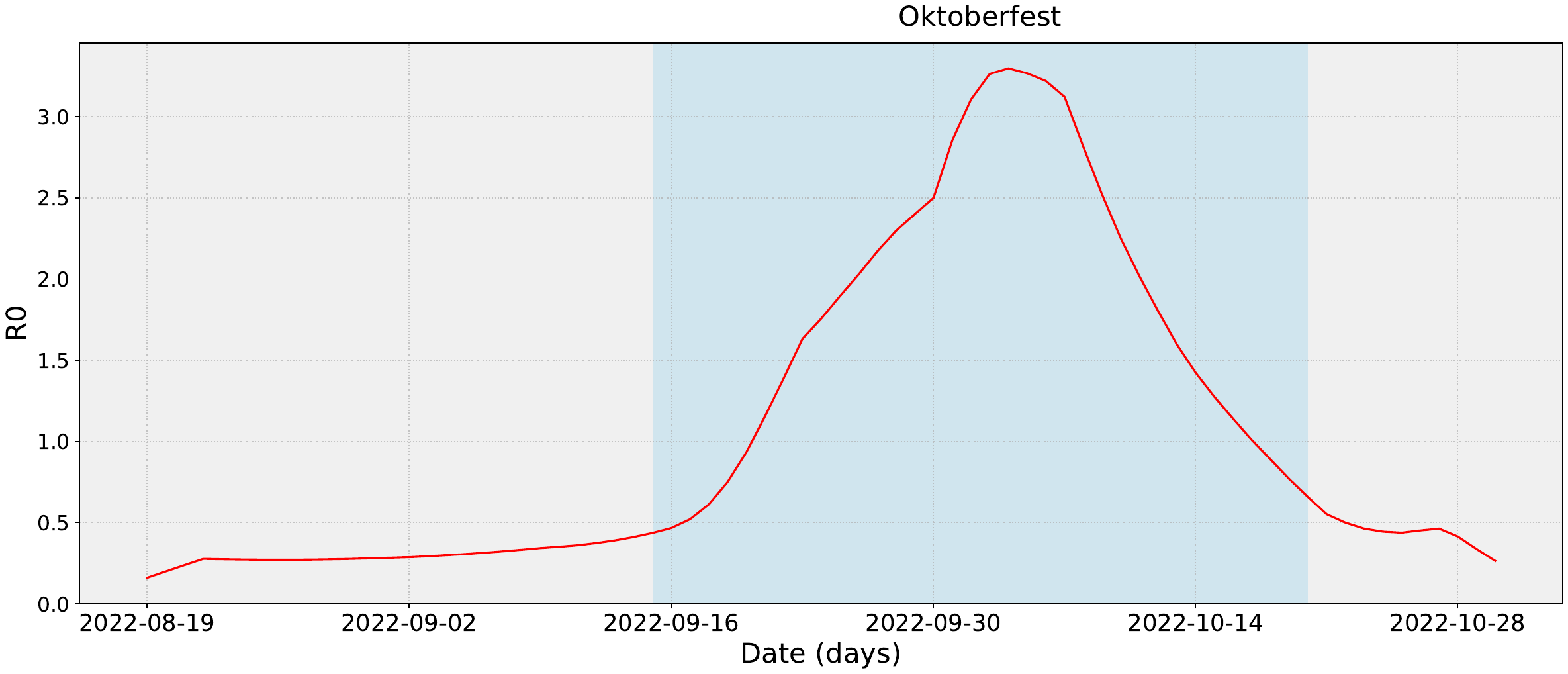}
    \captionsetup{font=normalsize}
    \caption{Visualization of $R_0$ curves in Bayern of Germany.}
    \label{fig:erng}
\end{figure}

Figures~\ref{fig:ernc} and~\ref{fig:erng} illustrate the trajectory of $R_0(t)$ for representative regions of China and Germany. In Figure~\ref{fig:ernc}, the trajectory for Shaanxisheng is shown, where shaded blue intervals correspond to key periods of increased human mobility and social interaction, such as the \emph{Back-to-school Season} and \emph{Golden Week}. During these intervals, $R_0(t)$ rises markedly, reflecting elevated transmission risks due to intensified population mixing. Once these events conclude, $R_0(t)$ declines again, signifying a restoration of lower transmission potential. \par 

Similarly, Figure~\ref{fig:erng} highlights the transient increase of $R_0(t)$ in the state of Bayern during Oktoberfest. The shaded blue area indicates the festival timeframe, during which $R_0(t)$ surges as crowd density and contact rates increase markedly. After the event, $R_0(t)$ steadily declines, aligning with a reduction in social mixing and return to stable patterns of lower mobility. These upward and downward shifts in $R_0(t)$, synchronized with the presence or absence of significant events, demonstrate the model’s ability to dynamically capture real-world social and mobility variations. \par

Interestingly, the maximal $R_0(t)$ values in Shaanxisheng are higher than those in Bayern during key events, reflecting an augmented transmission potential in China due to concentrated periods of increased mobility (e.g., \emph{Golden Week} and \emph{Back-to-school Season}). While Shaanxisheng has a larger total population, its population density is relatively similar to that of Bayern, suggesting that mobility patterns are the primary driver of these differences, though the potential influence of population density cannot be ruled out. However, the actual epidemic severity in Germany was greater, influenced by its more relaxed public health policies and a higher cumulative infection rate, which increased the effective susceptibility of the population \cite{Huber2020}. This discrepancy highlights the distinction between theoretical transmission potential captured by $R_0(t)$ and the realized epidemic impact shaped by regional policies and long-term dynamics.

The analysis of contact rate and effective reproduction number demonstrates that the Spatio-Contact SIR model not only provides accurate predictions but also enhances interpretability by uncovering the underlying spatio-temporal trends influencing population interactions. These results underscore the potential of our framework as a robust tool for epidemic forecasting and policy evaluation. \par

\section{Conclusion}
\label{sec:co}
Since the outbreak of COVID-19, accurate epidemic forecasting has emerged as a pivotal area of research. In this study, we introduce a novel framework that integrates mathematical modeling with neural networks to predict the spatio-temporal spread of infectious diseases. Our core framework is the Causal Spatiotemporal Graph Neural Network (CSTGNN), which incorporates the Spatio-Contact SIR model (SCSIR) to account for both intra-regional homogeneity and inter-regional heterogeneity. The SCSIR model serves as an embedded module, utilizing the contact rate parameter to capture human mobility patterns and effectively reflect the complexities of disease transmission. By embedding the SCSIR model into CSTGNN, we seamlessly merge epidemic mechanisms, causal inference, and data-driven techniques. This integration enhances the interpretability of the neural network and ensures that its training is guided by epidemiological knowledge. Furthermore, we incorporate temporal decomposition and graph learning modules to accurately model population movement and spatio-temporal dependence, enabling the framework to discern complex transmission dynamics and interactions. \par 

We validated the effectiveness of our proposed method using real-world data from China and Germany over the course of a year. This period encompasses a variety of temporal and spatial dynamics, including different seasons and major holidays and feasts in both countries. Comparative experiments against several baseline methods demonstrate that our framework consistently outperforms existing approaches in epidemic forecasting. Additionally, ablation studies highlight the significance of each model component, underscoring the framework’s robustness and precision in real-world applications. These results affirm our method’s capability to adeptly capture and leverage spatio-temporal patterns for reliable predictions. \par

Beyond forecasting accuracy, we delve into the analysis of learned epidemiological parameters, such as the contact rate matrix and the effective reproduction number, to better understand disease transmission dynamics in both countries. Our findings indicate that these parameters closely mirror real-world observations, effectively capturing the fundamental patterns of epidemic spread. Overall, our framework not only excels in predictive performance but also provides valuable, interpretable epidemiological insights. This dual advantage demonstrates its potential as a powerful tool for understanding and managing infectious diseases. Future research may extend this integration of mathematical models and neural networks to other domains, further harnessing the synergy between domain-specific knowledge and deep learning. \par     




\section*{Data availability}
Data will be made available on request.\par

\section*{Acknowledgments}
The present work was supported by multiple funding sources, including the XF-IJRC (S. Han), the ENABLE Project of HMWK (L. Stelz), the CUHK-Shenzhen university development fund under grant No.\ UDF01003041 and UDF03003041, and Shenzhen Peacock fund under No.\ 2023TC0179 (K. Zhou). We also thank Armin van de Venn for his enthusiastic support in the mathematical aspects of this work. Additionally, we sincerely appreciate Marcelo Netz-Marzola for his invaluable suggestions provided during the review process. The authors gratefully acknowledge the computational resources provided by the Center for Scientific Computing (CSC) and the Goethe-HLR cluster at Goethe University Frankfurt.\par 

\bibliographystyle{elsarticle-num} 
\bibliography{refs}

\end{document}